\documentclass{tlp}

\usepackage{amsmath}
\usepackage{graphicx}
\usepackage{multirow}
\usepackage{makecell}
\usepackage{tabularx}
\usepackage{hyperref}
\let\proof\relax

\usepackage{amsthm}
\usepackage{amssymb}
\usepackage{array}
\usepackage[dvipsnames]{xcolor}
\usepackage{listings}
\usepackage{pdflscape}
\usepackage{booktabs}
\usepackage{enumitem}
\usepackage{subcaption}
\newlength{\subcolumnwidth}

\newcommand{\nextsubcolumn}[1][]{%
  \cr\noalign{\hfill}
  \if\relax\detokenize{#1}\relax\else\hsize=#1\setlength{\subcolumnwidth}{\hsize}\fi
}

\newtheorem{example}{Example}
\theoremstyle{definition}
\newtheorem{definition}{Definition}

\begin{document}

\lstset{
  basicstyle=\ttfamily,
  columns=fullflexible,
  keepspaces=true,
}

\lefttitle{D. Fossemò et al.}

\jnlPage{1}{8}
\jnlDoiYr{2021}
\doival{10.1017/xxxxx}
\title{Explaining Neural Networks in Preference Learning: a Post-hoc Inductive Logic Programming Approach}
\shorttitle{Explaining Neural Networks with Post-hoc ILP in Preference Learning}

\begin{authgrp}
\author{\gn{Daniele Fossem\`{o}}}
\affiliation{Department of Information Engineering, Computer Science and Mathematics, University of L'Aquila, Italy }
\affiliation{\email{daniele.fossemo@graduate.univaq.it}}
\author{\gn{Filippo Mignosi}}
\affiliation{Department of Information Engineering, Computer Science and Mathematics, University of L'Aquila, Italy}
\affiliation{ICAR-CNR, Palermo, Italy}
\affiliation{\email{filippo.mignosi@univaq.it}}
\author{\gn{Giuseppe Placidi}}
\affiliation{Department of Life, Health and Environmental Sciences, University of L'Aquila, Italy}
\affiliation{\email{giuseppe.placidi@univaq.it}}
\author{\gn{Luca Raggioli}}
\affiliation{Department of Electrical Engineering and Information Technologies, University of Naples, Italy}
\affiliation{\email{luca.raggioli@unina.it}}
\author{\gn{Matteo Spezialetti}}
\affiliation{Department of Information Engineering, Computer Science and Mathematics, University of L'Aquila, Italy}
\affiliation{\email{matteo.spezialetti@univaq.it}}
\author{\gn{Fabio Aurelio D'Asaro}}
\affiliation{Ethos Group, Department of Human Sciences, University of Verona, Italy}
\affiliation{\email{fabioaurelio.dasaro@univr.it}}
\end{authgrp}

\history{\sub{xx xx xxxx;} \rev{xx xx xxxx;} \acc{xx xx xxxx}}

\maketitle

\begin{abstract}
    In {this} paper, we propose using Learning from Answer Sets to approximate black-box models, such as Neural Networks (NN), in the specific case of learning user preferences. We specifically explore the use of ILASP (Inductive Learning of Answer Set Programs) to approximate preference learning systems through weak constraints. We have created a dataset on user preferences over a set of recipes, which is used to train the NNs that we aim to approximate with ILASP. Our experiments {investigate} ILASP both as a global and a local approximator of the NNs. 
    These experiments address the challenge of approximating NNs working on {increasingly} high-dimensional feature spaces while achieving appropriate fidelity on the target model {and limiting the increase in} computational time. 
    To handle this challenge, we propose a preprocessing step that exploits Principal Component Analysis to reduce the dataset's dimensionality while keeping our explanations transparent.\\[1em]Under consideration for publication in Theory and Practice of Logic Programming (TPLP).
\end{abstract}


\section{Introduction}\label{section:introduction}
In recent years \textit{Machine Learning} (ML) algorithms have become pervasive in our daily life. This is thanks to the extensive research and results carried out over time on these methods, covering very diverse application areas, such as robotic and drone navigation~\citep{Mohsen2023AI}, medical imaging~\citep{Pranav2023MedIMag}, legal reasoning~\citep{Chuyue2024AiLaw}, {and many other domains}. However, its recent developments have highlighted the need to better understand and explain black-box ML models~\citep{Cath2018AiEtichs, Bostrom2014AiEtichs}. In fact, although these models prove very accurate in many application domains, their internal structures are exceedingly hard to understand and interpret, even for the developers themselves. For this reason, they are nowadays often referred to as \textit{black-box} (BB) models.
Recent regulations (such as the General Data Protection Regulation\footnote{\url{https://eur-lex.europa.eu/eli/reg/2016/679/oj}} and the recent AI Act\footnote{\url{https://www.europarl.europa.eu/doceo/document/TA-9-2024-0138_EN.pdf}}) have stressed the need for automated systems decisions to be explainable, especially when these may affect individuals in high-risk domains.

To address these concerns about BB, \textit{Explainable AI} (XAI) was proposed as a direction to create AI systems {whose} internal decision-making can be understood and appropriately trusted by end ``lay'' users~\citep{Gunning2019XAI}.
This field spotlights the need for \textit{transparent systems} that are able to provide a human-understandable rationale for their decision, as well as \textit{post-hoc methods}, which formulate explanations from BB outputs (see~\citep{gui18survey} for a survey of such methods).
An advantage of this latter approach is that they do not require to alter the black-box structure, thus not impacting the performance of these models. Transparent systems, on the other hand, often impose a trade-off between accuracy and explainability.
\paragraph{Motivation.}
{Preference learning represents a particularly relevant domain for explainability, as it directly models human choices and decision processes.}
{It tackles everyday decision processes of users in several contexts, ranging from recommendation systems to search tools.}
Additionally, this domain, formalized in Section \ref{subsection:Pref_Learn}, {naturally accommodates decision-making under uncertainty, for instance, by reflecting the noisy labels common in real-world tasks, thus providing an interesting and realistic application domain for XAI techniques.}
\paragraph{Problem statement.}
{In this work, we address the problem of explaining black-box models for preference learning by means of symbolic, interpretable theories.}
{We work in a post-hoc setting, where a neural network has already been trained to implement a preference model, and we seek to approximate its behaviour with logical theories learned by an Inductive Logic Programming (ILP) framework.}
{Following the formal perspective adopted later in Section~\ref{section:background}, an \emph{explanation} is a transparent (i.e., low-complexity) model that approximates the behaviour of a black box with high fidelity on a suitable region of the input space, in the sense of~\citep{ribeiro2016lime}.}

{We propose a post-hoc method to explain black-box models for preference learning using} Answer Set Programming (ASP)~\citep{calimeri2020asp} and the Inductive Logic Programming (ILP) framework ILASP~\citep{law2014inductive,law2015learning,law2018inductive}, building upon~\citep{d'asaro2020inductive,fossemò2022inductive}.
These systems can output logical theories that easily translate into natural language and, therefore, can be understood even by non-experts. 
For this reason, they are easier to check (when compared to non-transparent ML models) for the presence of biases and artifacts resulting from training on large collections of data (see, e.g.,~\citep{caliskan2017semantics,schramowski2017moral}).

Following~\citep{d'asaro2020inductive}, we deepen our exploration of ILASP as a method to analyze and produce theories from target black-boxes, which may prove useful in situations where the data the black-boxes were trained on is not available. 
This method may help identify systematic biases, such as, e.g., decisions influenced by prejudices or unfair principles.

In~\citep{fossemò2022inductive} we presented a preliminary implementation and workflow to use ILASP to approximate black-box models in the context of preference learning systems, by presenting methods to tackle practical limitations of ILASP when dealing with high-dimensional feature spaces.
{Building on that work, the present study extends both the experimental analysis and the evaluation framework. In particular, while the previous study focused on preliminary results limited to the global approximation of deep neural networks, this paper (i) refines the formal notion of approximation, distinguishing explicitly between local and global explanations, and (ii) provides a broader empirical evaluation of global approximation. Moreover, we include a new ground-truth-based metric that leverages the transparency of Inductive Logic Programming to assess the quality of the learned theories.}
\paragraph{Scope beyond preference learning.}
{It is worth noting that our method is not limited to preference learning. As discussed in Section~\ref{subsection:Pref_Learn}, our framework operates in the general context of ranking problems: we approximate a black-box model that orders items according to some (possibly latent) criterion.}
{Ranking tasks involving black-box models are not restricted to user preferences: examples include information retrieval, where documents are ranked by relevance~\citep{Severyn2015InfRetr}, and scheduling problems such as cluster scheduling~\citep{Li2022ClustShed}, where tasks are ordered to optimize system performance.}
{Our choice of preference learning is therefore motivated both by its intrinsic relevance and by its natural fit with weak constraints in ASP and ILASP.}

Our approach integrates Principal Component Analysis (PCA) to balance feature space reduction with the clarity and fidelity of the resulting theories to the intended models. To support this methodology, we developed a dataset capturing user preferences for recipes sourced from GialloZafferano,\footnote{\url{https://www.giallozafferano.it/}} a popular Italian culinary website. This dataset comprises two main sections: a recipes section, created using a crawler to collect the top recipes and their ingredients categorized into classes, and a user preferences dataset, assembled from surveys designed to representatively sample user tastes across the recipe spectrum. Both datasets underwent preprocessing, including normalization and standardization for recipes, and PCA and k-means clustering for user preferences.
We trained the black-box model on user preferences and approximated it with ILASP.
{The result of this approximation is a theory composed of weak constraints, which represents the user preference system deduced from the black-box model.}
The code to check and reproduce our experiments can be accessed through Github.\footnote{\url{https://github.com/DanieleF198/ILASP-as-post-hoc-method-in-a-preference-system}} 
\paragraph{Research questions.}
{The research questions addressed in this work are the following:}
\begin{itemize}
    \item {\textbf{Q1}: Is it possible to make a black-box model transparent (globally and locally) using an Inductive Logic Programming framework?}
    \item {\textbf{Q2}: What level of fidelity can be achieved with such an approach?}
    \item {\textbf{Q3}: Is the method scalable as the amount of data increases?}
\end{itemize}
\paragraph{Contributions.}
{The main contributions of this paper are as follows:}
\begin{enumerate}
    \item {We propose a framework to explain the decision-making process of black-box models in preference learning through Inductive Logic Programming, approximating the model both globally and locally;}
    \item {We address the exponential growth in execution time with respect to dataset size by introducing Principal Component Analysis in the pipeline;}
    \item {We introduce a novel ground-truth-based metric that leverages ILASP’s transparency to better assess the quality of learned logical theories;}
    \item {We present a curated dataset of recipes and user preferences, designed to evaluate explainable preference learning systems under realistic conditions.}
\end{enumerate}
\paragraph{Results and generality.}
{As shown in Section~\ref{subsection:results}, our proposed methods to tackle scalability issues are effective both in terms of execution time and fidelity. In fact, even in the worst cases, these methods manage to increase fidelity while reducing execution time. Furthermore, as described in Section~\ref{subsection:methods}, our approach is \emph{model-independent} in the sense that it treats the neural network as an oracle, querying it on suitably sampled instances, rather than relying on direct access to the original training data. In practice, the ILASP mode bias and the sampling procedures are instantiated from the feature representation of the available data, but the explanation pipeline itself does not require retraining or modifying the black-box model. Thus, the same methodology can be applied to contexts beyond preference learning by appropriately modifying the sampling technique, then processing the resulting instances with PCA (if needed), and adapting ILASP's mode bias to match the dataset features. Although this may look non-trivial, this pipeline can be automated, and we plan to do so in future work.}

{The remainder of this paper is structured as follows. Section~\ref{section:related_work} reviews the main contributions in the literature that are related to our work. Section~\ref{section:background} introduces the necessary background concepts, including preference learning, Answer Set Programming (ASP), ILASP, and the process of approximating a neural network through ILASP. Section~\ref{section:dataset} describes the datasets used in our experiments, namely the \textit{Recipes Dataset} and the \textit{Users Preferences Dataset}. Section~\ref{subsection:methods} details the proposed methodology, including the experimental setup for global and local approximation, the use of \textit{indirect} and \textit{direct} PCA, preliminary analyses, and the definition of truth scores. Section~\ref{subsection:results} presents and discusses the experimental results for both global and local approximations. Finally, Section~\ref{section:conclusions} concludes the paper and outlines possible directions for future work.}

\section{Related work}\label{section:related_work}
\paragraph{Explainable Artificial Intelligence.} {XAI is a research field that aims to make AI systems more understandable to humans~\citep{Adadi2018XAI}. This field of study has gained increasing attention in recent years, together with the success of machine learning and deep learning in many application areas. A key driver of this interest is the need to ensure transparency and accountability in AI systems, especially in critical domains such as healthcare, finance, and law enforcement. Recent taxonomies~\citep{Arrieta2020XAI,Weber2023Taxonomies,Fan2021Taxonomies} organise XAI methods along several dimensions. A useful distinction is between \emph{transparent-by-design}, \emph{model-specific}, and \emph{post-hoc} approaches. Transparent-by-design methods are simple enough to be interpretable (for example, decision trees or sparse linear models). Model-specific methods exploit the internal structure of complex models and try to reconstruct the reasoning that led to specific outputs. Post-hoc methods, by contrast, aim to explain the results of a black-box model without making strong assumptions about its internal structure~\citep{gui18survey}. Another widely used distinction is between \emph{global} and \emph{local} methods: global techniques seek to capture the overall decision logic of the black box, whereas local techniques focus on explaining the prediction for a particular input or a small neighbourhood of it. Our work fits into this landscape as a post-hoc, rule-based approach that can be instantiated both globally and locally, in the sense made precise in Section~\ref{subsection:explaining_black_boxes}.}
\paragraph{Previous work and ILP-based proposed approach}
This paper builds and extends upon~\citep{d'asaro2020inductive} and~\citep{fossemò2022inductive}. In the former, ILASP was applied as a post-hoc XAI method to local and global approximation of \textit{Support Vector Machines} (SVMs), a set of simple classical ML models that have often been used to benchmark explainable AI technologies (see e.g.~\citep{kamishima2003nantonac}). The present work addresses the need for enlarging the scope of black-boxes taken under consideration to more fashionable (and complex) ML models such as NNs.
In~\citep{fossemò2022inductive}, we continued the work with the creation of a dataset in order to have larger data and a more complex neural network to conduct studies.
{In that preliminary study, the focus was on global approximation of a neural network trained on a first version of the recipes dataset, and on showing that ILASP could, in principle, approximate the black box with weak-constraint theories.}
{In this paper, we present the final version of the dataset and a substantially extended set of experiments on both global and local approximation. We systematically investigate the behaviour of ILASP under variations of (i) the sampled dataset size, (ii) the parameters of the \emph{indirect} and \emph{direct} PCA methods, and (iii) the \texttt{\#maxp} ILASP hyperparameter. For the evaluation, in addition to the estimators described in Section~\ref{subsection:hyperparameters}, such as the classical fidelity indicator~\citep{gui18survey}, we also consider execution time, the length of the returned theories, and ground-truth scores based on the weak-constraint definition. This more extensive analysis clarifies in which regimes ILASP can provide faithful and compact explanations of neural preference models.}

{Deep neural network architectures currently represent the state of the art in preference learning and recommendation, as they effectively capture complex patterns of user: item relevance and ranking relations. For this reason, a DNN model was adopted in our experiments as the black-box to be approximated. Representative examples of such architectures include DNNs combined with matrix factorisation~\citep{zhu2020dnnPrefLearn}, dual adversarial networks~\citep{zhang2021dnnPrefLearn}, cross-domain DNNs~\citep{Hong2020dnnPrefLearn}, and aesthetic cross-domain networks~\citep{Liu2022dnnPrefLearn}. Our goal is not to propose a new recommendation architecture, but to study how an ILP-based surrogate can approximate and explain a reasonably powerful neural preference model. The framework proposed in this paper can in principle be applied on top of more sophisticated recommendation architectures, providing a symbolic approximation layer that makes their learned preference structures explicit and interpretable.}
\paragraph{Alternative ILP frameworks.} Our work is not the only one using ILP to approximate black-boxes. For instance,~\citep{rabold2018explaining,rabold2020enriching} and~\citep{shakerin2019induction} use the Prolog-based system Aleph~\citep{srinivasan2004aleph}. We use ILASP instead, which was shown to outperform Aleph as well as decision trees~\citep{Ville2013DT} when applied to preference learning dataset~{\citep{law2018thesis}}.
Even if not used in similar task, a mention must be made to FastLAS~\citep{Law2020FastLas}. FastLAS is an ILP system based on the ILASP framework that solves a restricted version of the context-dependent learning from answer set task. This method received attention thanks to its faster performance on large-scale ILP benchmarks. However, FastLAS in its current form does not yet support weak constraint learning using ordering examples, which is central to our use case. 
{In our setting, the ability to learn weak constraints and explicit rules is not only a technical requirement but also a central ingredient of transparency: the learned weak constraints make explicit the trade-offs and relative penalties that drive the underlying preference model. This justifies the use of ILASP, despite its higher computational cost compared to FastLAS, because it offers the expressivity needed to capture fine-grained preference structures and to present them in an interpretable, rule-based form.}
This limitation underscores the novelty and technical challenge of our work, which leverages ILASP’s unique capabilities for learning preferences in a logical framework.
\paragraph{Neurosymbolic ILP approaches.} {Integrated neurosymbolic methods have recently received considerable attention for their ability to combine neural networks, which handle non-symbolic data such as images, audio, and raw text, with symbolic reasoning. For instance, Collevati et al.~\citep{Collevati2024NeuroSymbolic} propose a neurosymbolic pipeline for slice discovery that extracts logical rules describing model behaviours and then uses those rules to generate new data to amend the underlying classifier. Other lines of work pursue similar goals with different design choices, such as Hillerström et al.~\citep{Hillerström2024NeuroSymbolic}, which aims to handle probabilistic background knowledge by using neural components to predict the probability that a predicate holds (for example, in an object-detection scenario). Our method, by contrast, has a different objective: it is intended purely as a \textit{post-hoc} explanation mechanism for black-box models, rather than as a way to improve their predictive performance or correct them through additional training. Nonetheless, the integration of neurosymbolic components could be explored in future work to extend our approach to non-symbolic input domains.}
\paragraph{Non-ILP approaches to black-box explanation.} ILP is not the only possible approach to explaining black-box systems: for instance, Exceptional Preferences Mining~\citep{rebelo2016exceptional} and LIME~\citep{ribeiro2016lime} are other techniques whose relationship to ILP and ILASP may be the subject of future work.
Specifically,~\citep{Brunot2022LimePref} proposes LIRE on the MovieLens~\citep{Harper2015MovieLens} dataset, a local surrogate model tailored for recommender system which extends LIME. Another work in the same domain is presented in~\citep{Hu2022ShapPref}, where Pref-SHAP is introduces, a Shapely value-based model explanation framework for pairwise comparison data, applied to a variety of synthetic and real-world dataset. {These methods provide feature-attribution style explanations around specific predictions of recommendation models and have been influential in the XAI literature for recommender systems.} Despite these methods are promising, {they fundamentally differ from our approach. LIRE focuses on rating prediction rather than on learning global, symbolic preference theories, and both LIRE and Pref-SHAP provide local importance scores but do not aim at inducing an explicit, human-readable rule set that captures the overall preference structure.} Indeed, LIRE focuses on classification rather than preference learning, and both LIRE and Pref-SHAP do not account for uncertainty. Furthermore, our method allows to get interpretable results which can be easily translated in natural language so that are easy to understand even to non-experts.
We previously mentioned that our technique is a \textit{post-hoc} one, namely it applies to any black-box whose training was performed before the explanation procedure takes place. Other powerful techniques can be deployed if we relax this constraint and perform model-specific operations throughout the training phase, as in~\citep{ferreira2022looking} that is specific to NNs and requires some additional labeling effort. These systems are potentially more accurate than ours. Nonetheless, they are not post-hoc and thus are less generally applicable than our approach.

\section{Background}\label{section:background}

\subsection{Preference Learning}
Preference Learning~\citep{furnkranz2010preference,gurrieri2012advances,zhou2014taxonomy} is a subfield of Artificial Intelligence (AI) whose purpose is profiling user preferences. This field deals with several scenarios: for instance, a search engine that must return the results according to the user past searches; a streaming platform that needs to order videos according to the user's preferences. In both examples, Preference Learning aims to build a system that collects user data and returns the results that are likely to be preferred by the user.

{The problem we have informally described here is to order a set of elements based on user preferences, i.e., a \emph{ranking} problem. Among the various instantiations of ranking, we mainly focus on \emph{label ranking}.}
Given an instance space \textit{\textbf{X}} and a set of labels $L = \{\lambda_{1}, \lambda_{2}, ..., \lambda_{k}\}$,  an ordering for an instance \textit{\textbf{x}} can be described as ordering of the labels by means of a relation $\succ_{\textit{\textbf{x}}}$ such that $\lambda_{\textit{i}}\succ_{\textit{\textbf{x}}}\lambda_{\textit{j}}$ indicates that $\lambda_{\textit{i}}$ is preferred over $\lambda_{\textit{j}}$ by instance \textbf{\textit{x}}.
So the ordering induced by $\succ_{\textit{\textbf{x}}}$ can be identified by a permutation $\pi_{\textit{\textbf{x}}}$ such that $\pi_{\textit{\textbf{x}}}(\textit{i}) = \pi_{\textit{\textbf{x}}}(\lambda_{i})$ indicates the position of the label $\lambda_{i}$ in the ordering. 
So, a complete ordering of the labels in \textit{L} can be described as following:
\begin{equation}
        \lambda_{\pi_{\textit{\textbf{x}}}^{-1}(1)} \succ_{\textit{\textbf{x}}} \lambda_{\pi_{\textit{\textbf{x}}}^{-1}(2)} \succ_{\textit{\textbf{x}}} ... \succ_{\textit{\textbf{x}}} \lambda_{\pi_{\textit{\textbf{x}}}^{-1}(\textit{k})}
    \label{eq:prefPerm}
\end{equation}
\noindent where $\pi_{\textit{\textbf{x}}}^{-1}(\textit{j})$ indicates the index of the label which occupies the \textit{j}-th position in the ordering~\citep{zhou2014taxonomy}.
An ordering of elements can be decomposed in the form of \textit{pairwise comparison}.
The idea behind this decomposition is that the ordering of a set of elements can be structured as a succession of consecutive questions ``do you prefer $\lambda_{\textit{i}}$ over $\lambda_{\textit{j}}$?'' (with \textit{i} $\ne$ \textit{j}) in which the answers could be either ``yes'' or ``no''.
Thus, knowing the position of each element, we can define the preference ordering as follows:
\begin{equation}
    \textit{f}(\lambda_{\textit{i}}, \lambda_{\textit{j}})=\begin{cases}
        \lambda_{\textit{i}}\succ_{\textit{\textbf{x}}}\lambda_{\textit{j}}, & \text{if $\lambda_{\textit{i}}$ preceds $\lambda_{\textit{j}}$ in the ordering}\\
        \lambda_{\textit{j}}\succ_{\textit{\textbf{x}}}\lambda_{\textit{i}}, & \text{otherwise}
    \end{cases}
\end{equation}
\noindent where the number of pairs $(\lambda_{\textit{i}}, \lambda_{\textit{j}})$, if \textit{k} indicates the number of elements in the collection, is equal to $\textit{k}(\textit{k}-1)/2$.
This decomposition allows us to decompose the original problem into a set of simpler sub-problems, which is useful from a machine learning perspective~\citep{furnkranz2010preference}.
In our work we deal with the case where the relationship between any two elements may be unknown or uncertain, i.e., for an instance \textit{\textbf{x}} and indices $i$, $j$ we are unsure as to whether $\lambda_i \succ_{\textit{\textbf{x}}} \lambda_j$ or $\lambda_j \succ_{\textit{\textbf{x}}} \lambda_i$.
For pairs of elements for which the preference relationship is unknown, we label them as \textit{uncertain} pairs (denoted with $\lambda_i \sim_{\textit{\textbf{x}}} \lambda_j$). This makes our problem a \textit{ternary} classification problem.

\begin{equation}
    p_x(\lambda_i, \lambda_j)=\begin{cases}
    1 & \text{if }\lambda_{{i}}\succ_{{x}}\lambda_{{j}}\\
    0 & \text{if } \lambda_{{i}}\sim_{{x}}\lambda_{{j}}\\
        -1 & \text{if } \lambda_{{j}}\succ_{{x}}\lambda_{{i}}
    \end{cases}
\end{equation}

{In the rest of the paper, we will often view $p_x(\lambda_i,\lambda_j)$ as the (symbolic or neural) output of a pairwise comparison model $f_x$ for user $\mathbf{x}$, which will later be approximated by ILASP theories.}

\label{subsection:Pref_Learn}
\subsection{Answer Set Programming}
\textit{Answer Set Programming}~\citep{calimeri2020asp} is a declarative logic programming language based on the stable model semantics, whose output is a model of the input theory called \textit{stable model} or \textit{answer set}.
This language is at the base of our work thanks to the possibility to define \textit{weak constraints}.
In fact, weak constraints give us a suitable and natural way to represent the preference relation $\succ_{\textit{\textbf{x}}}$ in ASP, inducing a (partial) ordering over the stable models of a theory.
Given \textit{n}, \textit{m} $\ge$ 0, a weak constraint has the form:
\begin{equation}
        \verb|:~ b1, ..., bn [w@l, t1, ..., tm]|
\end{equation}
where \verb|b1|, $\dots$, \verb|bn| are \textit{literals}, \textit{w} is a \textit{weight} associated with the constraint, \textit{l} is a \textit{priority level}, and \verb|t1|, $\dots$, \verb|tm| are used to handle independence among weak constraints.
We are not going to discuss this syntax here, but the interested reader can find the full syntax and semantics in~\citep{calimeri2020asp}.
The important difference with hard constraints is that while these affect the answer set of a theory by excluding some solution at prior, weak constraint induce a preference relation among them, exactly as we want to naturally reflect a user preference system.
To describe this preference relation, we need to introduce the notion of \textit{cost} of an answer set at some priority level \textit{l}.
This is defined as the sum of weights at priority level \textit{l} for all the weak constraints such that their bodies are satisfied by the answer set.
\begin{example}
    Consider the ASP program P consisting of the following axioms and weak constraint:
    \begin{verbatim}
        p(a). p(b). p(c).
        0{ q(X) }1 :- p(X).
        :~ q(a). [1@2, a]
        :~ q(b). [3@1, b]
        :~ q(c). [-1@2, c]
    \end{verbatim}
\end{example}

\noindent This theory has 8 answer sets, namely: $ \{p(a), $ $p(b),$ $p(c)\}$, $\{p(a),$ $p(b),$ $p(c),$ $q(b)\}$, $\{ p(a),$ $p(b),$ $p(c),$ $q(c)\}$, $\{p(a),$ $p(b),$ $p(c),$ $q(b),$ $q(c)\}$, $\{p(a),$ $p(b),$ $p(c),$ $q(a)\}$, $\{ p(a),$ $p(b),$ $p(c),$ $q(a),$ $q(c)\}$, $\{ p(a),$ $p(b),$ $p(c),$ $q(a),$ $q(b)\}$ and $\{p(a),$ $p(b),$ $p(c),$ $q(a),$ $q(b),$ $q(c)\}$.
{These eight answer sets are obtained by \emph{ignoring} the weak constraints, that is, by computing all stable models of the hard part of the program and then evaluating their costs afterwards. Weak constraints do not change which answer sets exist, but they induce a preference ordering over them. When weak constraints are taken into account, answer sets are ranked lexicographically by their total cost at each priority level: one first minimises the sum of weights at level $1$, and, among those tied, minimises the sum at level $2$, and so on. Under this criterion, the optimal (i.e., most preferred) answer set is $\{ p(a),$ $p(b),$ $p(c),$ $q(c)\}$, which has total cost $0$ at level $1$ and $-1$ at level $2$ (the lowest at each level).}
{Moreover, consider $A$ =  $\{ p(a), p(b), p(c), q(a), q(b), q(c) \}$ and $B$ = $\{ p(a), p(b), p(c), q(a), q(b) \}$. Then $A$ is preferred over $B$ with respect to weak constraints, since even though both have cost $3$ at level $1$, $A$ has a lower cost than $B$ at level $2$ (i.e., $0$ vs $1$).}
In general, we say that an answer set $A$ is preferred to an answer set $B$ (according to theory $T$) if the cost of $A$ is strictly smaller than that of $B$ at the highest level for which their costs differ.
\label{subsection:ASP}

\subsection{ILASP}
ILASP (short for \textit{Inductive Learning of Answer Set Programs})~\citep{law2014inductive,law2015learning,law2018inductive} is an Inductive Logic Programming language which can learn ASP programs, including weak constraints.
This means that we can represent preferences using weak constraints in ASP and make ILASP learn them.
{Moreover, ILASP's output can be expressed in natural language, although this translation may require some effort and may vary from case to case. Nonetheless, the structured and human-readable form of ILASP rules supports its effectiveness as an XAI method by providing clear and transparent explanations that can be made accessible to non-expert end users.}
{In the remainder of this section, we focus on the abstract ILASP formalisation rather than on concrete syntax. Full syntactic details and complete code listings for the examples are reported in \ref{appendixA}.}

The three main components of ILASP are the \textit{background knowledge}, the \textit{language bias} and the \textit{examples}.
The background knowledge $B$ is an ASP program and represents the prior knowledge ILASP has prior to rule induction.
The language bias $L$ (often called a \textit{mode bias}) is used to specify the so-called \textit{hypotesis space} or \textit{search space}, which is the set of rules that ILASP can learn.
Language bias can be defined both explicitly and through the \textit{mode declarations}.
Since in many situations it is not practical to define the language bias explicitly, it is preferable to use mode declarations, which are ground atoms whose arguments are the so-called \textit{placeholders}.
{Intuitively, mode declarations specify which predicates may appear in the head or body of learned rules, which arguments are variables or constants of given types, and which atoms may occur inside weak constraints. In this way, $L$ compactly characterises the space $S_L$ of candidate hypotheses without committing to a particular syntax.}

{Lastly, the examples are observations guiding the learning process. Examples are divided into positive and negative examples (denoted $E^+$ and $E^-$, respectively), indicating which situations should or should not be entailed by the target theory.}
However, using only the examples, it is not possible to make ILASP learn the weak constraints, as this subdivision does not give any information on how learnable answer sets should be ordered.
For this reason, ILASP allows a second type of example called an \textit{ordering example}, which serves to define a preference ordering over a set of answer sets.
These ordering examples are crucial in our work, since they give us the point of contact between preference learning, ASP and ILP systems.
{Cautious and brave ordering examples constrain, respectively, all or at least one of the answer sets extending one example to be preferred to those extending another, according to the weak-constraint optimisation. In our setting, brave orderings are used to encode pairwise user (or neural) preferences between items.}

Finally, we can define what an ILASP task is:
\begin{definition}
    An ILP task is a tuple $T=\langle B, S_{L}, E^{+}, E^{-} \rangle$. 
    A hypothesis $H$ is an inductive solution of $T$ (written $H \in ILP_{LAS}(T)$) if and only if:
    \begin{enumerate}[leftmargin=*]
        \item \textit{$H \subseteq S_{L}$}
        \item \textit{$\forall e^{+} \in E^{+} \exists A \in AS(B \cup H)$ such that $A$ extends $e^{+}$}
        \item \textit{$\forall e^{-} \in E^{-} \nexists A \in AS(B \cup H)$ such that $A$ extends $e^{-}$}
    \end{enumerate}
    where $S_L$ is the search space induced by the language bias $L$ {and $AS(\cdot)$ is the set of all answer sets of an ASP program $(\cdot)$.}
\end{definition}

\begin{example}
    Consider the following code:

    \begin{verbatim}
        #modeo(1, value(const(val), var(val))).
        #modeo(1, category(const(mg)), (positive)).
        #weight(val).
        #weight(1).
        #weight(-1).
        #constant(mg, 1).
        #constant(mg, 2).
        #constant(val, p).
    
        #pos(item0,{},{},{category(2). value(p, 1).}).
        #pos(item1,{},{},{category(1). value(p, 2).}).
        #pos(item2,{},{},{category(2). value(p, 3).}).
        
        #brave_ordering(o0@2,item2,item1,>).
        #brave_ordering(o1@2,item2,item0,>).
        #brave_ordering(o2@1,item1,item0,<).
    \end{verbatim}
\end{example}
Here the  first block of code is used to define the language bias $L$ which defines the search space. 
the declaration \verb|#modeo(1, value(const(val), var(val)))| puts the following constraints in the search space:
\begin{lstlisting}[caption=constraints generated by the first line in Example 2]
    :~ value(p,V1).[1@2, V1]
    :~ value(p,V1).[-1@2, V1]
    :~ value(p,V1).[V1@2, V1]
    :~ value(p,V1).[-V1@2, V1]
    :~ value(p,V1).[V1@1, V1]
    :~ value(p,V1).[-1@1, V1]
    :~ value(p,V1).[1@1, V1]
    :~ value(p,V1).[-V1@1, V1]
\end{lstlisting}

\textit{Brave orderings} allow us to express a preference relationship between items. In this example, \verb|item2| is not preferred to \verb|item1| and \verb|item0|, while \verb|item1| is preferred to \verb|item0|.
Note that in the first two \verb|#brave_ordering| declarations the penalty is set to $2$, and so ILASP will give a penalty score equal to $2$ to those theories that do not cover those orderings.
Jointly with the hypothesis space defined by the \verb|#modeo| declaration, the program will learn a theory consisting of weak constraints expressing the order preferences exploited by brave orderings.
In this case the returned theory is:
\begin{lstlisting}[caption=Theory returned by Example 2]
    :~ value(p,V1).[V1@1, V1]
\end{lstlisting}
meaning that, given the defined cost, the higher the \verb|value| of \verb|p|, the better.
{Abstractly, this example illustrates a typical ILASP setup: background knowledge encodes the space of possible items, the language bias specifies which weak-constraint patterns can be used to express preferences, and brave orderings encode training preferences that guide the search towards a hypothesis $H$ capturing a monotonic relationship between an attribute (\texttt{value/2}) and preference. The concrete syntax above is only one possible realisation of this abstract schema.}

\label{par:ILASP}

{\subsection{Explaining black boxes}\label{subsection:explaining_black_boxes}
In this section we clarify what it means for a black-box model (such as a Neural Network) to be \emph{approximated} and \emph{explained}. Following \cite{ribeiro2016lime}, an explanation is a \emph{transparent} (i.e., low-complexity) model that approximates the behaviour of a black box with high fidelity on a suitable region of the input space.\footnote{{In \cite{ribeiro2016lime}, this is instantiated by minimizing, for each prediction, a locality-aware unfaithfulness term plus a complexity penalty. Our presentation abstracts this setup to generic regions of the input space and generic error and complexity measures.}}
\subsubsection{Preliminaries}
Formally, let $f : X \rightarrow Y$ be the function computed by a fixed black-box model, and let $\mathcal{T}$ be a class of candidate white-box models with the same input space $X$ (for us, typically ILASP theories). 
We assume:
\begin{itemize}
  \item an \emph{error} (or \emph{unfaithfulness}) measure
  \[
    L(T,f,A) \in \mathbb{R}_{\ge 0}
  \]
  that quantifies how poorly $T \in \mathcal{T}$ approximates $f$ on a non-empty region $A \subseteq X$;
  \item a \emph{complexity} measure
  \[
    \Omega : \mathcal{T} \rightarrow \mathbb{R}_{\ge 0}
  \]
  that penalises non-interpretable models (e.g., number of rules, depth of a tree, number of non-zero weights). Small values of $\Omega(T)$ correspond to \emph{transparent} or \emph{low-complexity} models.
\end{itemize}
Intuitively, \cite{ribeiro2016lime} first isolates fidelity to the black box, and only then adds a complexity constraint: a model that nearly minimises $L$ is a good \emph{approximation} of $f$; among those, the models with low $\Omega$ are taken as \emph{explanations}. This leads to the following.
\begin{definition}[Approximation on a region]
Fix a region $A \subseteq X$ and a tolerance $\varepsilon \ge 0$. A model $T \in \mathcal{T}$ is an \emph{approximation} of $f$ on $A$ if
\[
  L(T,f,A) \;\le\; \inf_{T' \in \mathcal{T}} L(T',f,A) + \varepsilon.
\]
That is, $T$ almost minimises the error of approximating $f$ on $A$.
\end{definition}
\begin{definition}[Explanation on a region]
Fix in addition a complexity threshold $\kappa \ge 0$. A model $T \in \mathcal{T}$ is an \emph{explanation} of $f$ on $A$ if:
\begin{enumerate}[leftmargin=*]
  \item $T$ is an approximation of $f$ on $A$ (as above), and
  \item $\Omega(T) \le \kappa$ (i.e., $T$ is transparent).
\end{enumerate}
Intuitively, an explanation is a low-complexity model that approximates $f$ with high fidelity on $A$.
\end{definition}
\subsubsection{Local and global explanations}
We now specialise the choice of the region $A$ to distinguish global and local explanations, again in the spirit of \cite{ribeiro2016lime}.
\begin{definition}[Global explanation]
Let $X_{\mathrm{rel}} \subseteq X$ be a task-relevant region of the input space (for example, the subset of $X$ on which $f$ is intended to be used). A model $T \in \mathcal{T}$ is a \emph{global explanation} of $f$ (relative to $X_{\mathrm{rel}}$) if it is an explanation of $f$ on $X_{\mathrm{rel}}$, i.e., it approximates $f$ well on $X_{\mathrm{rel}}$ and has low complexity.
\end{definition}
To capture locality, we endow $X$ with a metric $d$ that induces neighbourhoods around each input.
\begin{definition}[Local explanation]
Let $(X,d)$ be a metric space, and let $x^\star \in X$ be a specific query instance. For $r>0$, write
\[
  B_r(x^\star) \;=\; \{\, x \in X \mid d(x,x^\star) \le r \,\}
\]
for the closed ball of radius $r$ around $x^\star$. A model $T_{x^\star} \in \mathcal{T}$ is a \emph{local explanation} of $f$ at $x^\star$ if it is an explanation of $f$ on $B_r(x^\star)$, i.e., it achieves low error $L(T_{x^\star},f,B_r(x^\star))$ while having low complexity $\Omega(T_{x^\star})$.
\end{definition}}

\subsubsection{ILASP approximation of Neural Networks}

With respect to the notation introduced in Section~\ref{subsection:Pref_Learn}, let $\mathbf{x}$ be a user, let $L = \{\lambda_1,\lambda_2,\dots,\lambda_k\}$ be the set of elements that we aim to order according to the preference system of $\mathbf{x}$, and let $T_{\mathbf{x}}$ be the ILASP theory learned from data about $\mathbf{x}$. Then, we model the pairwise preference relations $\succ_{\mathbf{x}}$ and $\sim_{\mathbf{x}}$ induced by $T_{\mathbf{x}}$ as follows:
\begin{equation}
    T_{\mathbf{x}}(\lambda_i,\lambda_j) = 
    \begin{cases}
        \lambda_i \succ_{\mathbf{x}} \lambda_j, & \text{if $\lambda_i$ has lower cost than $\lambda_j$ w.r.t.\ $T_{\mathbf{x}}$},\\[2pt]
        \lambda_j \succ_{\mathbf{x}} \lambda_i, & \text{if $\lambda_j$ has lower cost than $\lambda_i$ w.r.t.\ $T_{\mathbf{x}}$},\\[2pt]
        \lambda_i \sim_{\mathbf{x}} \lambda_j,   & \text{if $\lambda_i$ and $\lambda_j$ have the same cost w.r.t.\ $T_{\mathbf{x}}$}.
    \end{cases}
    \label{eq:Tclass}
\end{equation}
Thus, starting from the pairwise comparison problem, we can establish (partial) orderings of the elements in $L$ by using the ranking formulation introduced in~\eqref{eq:prefPerm}.

In practice, $T_{\mathbf{x}}$ is learned from a finite \emph{ground-truth} set of already ordered pairs for user $\mathbf{x}$. Following the code example in \ref{AppendixD}, the ILASP \emph{brave ordering} encodes this ground truth, and the returned theory $T_{\mathbf{x}}$ is the corresponding preference system learned by ILASP.\footnote{Note that there is a discrepancy in notation between ILASP and preference learning. In ILASP, \emph{brave orderings} use $>$ to state that the \emph{second} item is preferred over the first, whereas in preference learning the symbol $\succ$ denotes that the \emph{first} item is preferred over the second. Although this may be slightly confusing, we keep this convention to remain consistent with the respective literatures.}

From the point of view of the general framework introduced above, we can regard the (unknown) user preference system as a target function
\[
  f_{\mathbf{x}} : X \rightarrow Y,
\]
where $X$ is the space of ordered pairs (e.g.\ feature representations of $(\lambda_i,\lambda_j)$) and $Y$ encodes the outcome of the pairwise comparison (e.g.\ ``$\lambda_i$ preferred'', ``$\lambda_j$ preferred'', or ``tie''). The ILASP theory $T_{\mathbf{x}}$ is an interpretable model in a hypothesis class $\mathcal{T}$ that approximates $f_{\mathbf{x}}$ on some region $A \subseteq X$ in the sense that it nearly minimises the error $L(T_{\mathbf{x}},f_{\mathbf{x}},A)$ introduced in the previous subsection. Since we only observe a finite ground-truth set of pairs, we cannot determine whether $T_{\mathbf{x}}$ perfectly aligns with the true preference system of $\mathbf{x}$ on unseen pairs; we therefore treat $T_{\mathbf{x}}$ as an \emph{approximation} of $f_{\mathbf{x}}$ on the relevant region $A$.

In the black-box explanation setting considered in this paper, the role of the target function is played by a neural network $f : X \rightarrow Y$ trained to model the preference system of user $\mathbf{x}$. Instead of using human-labelled pairs as ground truth, we use the predictions of $f$ as labels for training ILASP. The resulting theory, which we denote by $T_{NN}$, is then an interpretable surrogate that approximates the neural network on a region $A \subseteq X$ by (approximately) minimising $L(T_{NN},f,A)$. Whenever $T_{NN}$ achieves low error and has low complexity (e.g.\ few rules, shallow structure), it qualifies as an \emph{explanation} of the neural network: it is a transparent model that closely approximates the black box on $A$.

Following standard terminology on explanation properties~\cite{gui18survey}, we use \emph{fidelity} to describe how closely such an interpretable model approximates the neural network. In our pairwise setting, and in line with~\eqref{eq:Tclass}, we estimate fidelity by comparing the outcome $T(\lambda_i,\lambda_j)$ with the prediction $NN([\lambda_i,\lambda_j])$ over a sample of pairs, and we derive quantitative scores from these agreements and disagreements. The specific fidelity scores used in our experiments are described in Sections~\ref{subsection:hyperparameters} and~\ref{subsection:gt-scores}, while the neural network architecture is reported in \ref{AppendixB}.

In our specific setting, the black-box model $f$ is a neural network defined on the feature space $F_{NN}$ of pairs $(\lambda_i,\lambda_j)$, so that $X = F_{NN}$, each instance $x^\star \in X$ corresponds to one such pair, and the surrogates $T$ (for global approximation) and $T_{x^\star}$ (for local approximation) are instantiated as symbolic ILASP theories learned from $f$ according to the general local/global definitions given above.
\color{Black}

\label{subsection:approximatingNN}

\section{The Dataset}
In this section, we discuss the dataset we built to carry out our experiments on. The interested reader can find the dataset online.\footnote{\url{https://doi.org/10.5281/zenodo.7135196}} The dataset concerns users' gastronomic preferences. We collected recipes, their main features, and users’ preferences over these recipes.
{The public availability of this resource is itself one of the contributions of this work: the dataset contains $100$ curated recipes, $277$ distinct ingredients organised into $36$ ingredient classes and $12$ meta-classes, and explicit preference data from $54$ users. For each user we derive $210$ labelled pairwise comparisons, yielding $54 \times 210 = 11\,340$ annotated pairs in total. This combination of rich structured features and dense pairwise preference information is, to the best of our knowledge, not available in existing open datasets for food recommendation and preference learning.}
{We carefully considered existing benchmarks for preference learning and recommendation, both food-oriented (e.g., sushi preference datasets~\citep{kamishima2003nantonac}) and generic recommendation datasets such as MovieLens~\citep{Harper2015MovieLens}. However, these datasets either lack fine-grained, interpretable ingredient-level features or do not expose a structured taxonomy of attributes that can be naturally mapped to logical predicates and weak constraints. In this work, we explicitly leverage different levels of granularity in the feature space (ingredient, class, meta-class), which are crucial both for ILASP-based rule learning and for human-understandable explanations. For this reason, we rely on the dataset presented in~\citep{fossemò2022inductive} and extend it here as a reusable resource that can serve as a benchmark for interpretable preference learning, rule extraction, and XAI studies.}

\label{section:dataset}
\subsection{Recipes Dataset}
This dataset was {automatically collected by means of} a crawler that scraped data from the popular Italian recipe database GialloZafferano {and then manually verified and processed to ensure data quality}.\footnote{\url{https://www.giallozafferano.it/}}
From the $4000$ recipes in the database, we chose the $100$ most voted ones and gathered $277$ ingredients, which we grouped into $36$ \textit{classes} and $12$ \textit{meta-classes}, obtaining three levels of granularity for the ingredients. For example, the ingredient ``spaghetti'' belongs to the class ``dry pasta'' which in turn belongs to the meta-class ``pasta''.
In \ref{AppendixC} we report ingredients, classes, and meta-classes.

In order to deal with the challenge of balancing specificity and dataset size, we organized the ingredients into a hierarchical classification. With 277 ingredients, it would be impractical for participants to evaluate each one in a single survey. To address this, we utilized meta-classes and classes instead of individual elements.
On the other hand, since ILASP struggles with high-dimensional feature spaces, we performed ILASP experiments with the $12$ ingredients meta-classes, which makes the execution time computationally feasible in our application.
The dataset structure is outlined in Table \ref{table:recipesDatasetFeatures}. Before analysis, each recipe's ingredient vector and preparation vector were normalized, and
each numerical feature was standardized.
We manually assigned a level of importance to 
each recipe and each of the ingredients associated with it. To ensure the importance of ingredients is represented consistently at different granularities, we aggregate the importance values for each class. This means that for each class, we add up the importance values of the ingredients that belong to that class (or meta-class). For example, if a recipe includes ``pancetta'' (bacon) with an importance value equal to $4$, and ``prosciutto'' (ham) with an importance value equal to $2$, both of which are part of the ``pork meat'' class, then the ``pork meat'' class will have an importance value equal to $6$.

\begin{table}[h!]
	\centering
    \caption{Recipes Dataset features}
    \resizebox{0.9\textwidth}{!}{
        {\tablefont\begin{tabularx}{\textwidth}{@{\extracolsep{\fill}} cp{0.75\textwidth}}
            \topline
            \textbf{Feature} & \makecell{\textbf{Description}} \\ 
            \topline
            \textbf{ID} & \makecell{A unique integer assigned to the recipe}
            \midline
            \textbf{Name} & \makecell{The name of the recipe}
            \midline
            \textbf{Category} & \makecell{An integer from $1$ to $5$, where:\\ $1=$ starter; $2=$ complete meal; $3=$  first course; $4=$ second course;\\ $5=$ savory cake}
            \midline
            \textbf{Cost} & \makecell{An integer from 1 to 5, where:\\ $1 = $  very low; $2 = $ low; $3 = $ medium; $4 = $ high; $5 = $ very high}
            \midline
            \textbf{Difficulty} & \makecell{An integer from $1$ to $4$, where:\\ $1 = $  very easy; $2 = $ easy; $3 = $ medium; $4 = $ difficult}
            \midline
            \textbf{Preparation time} & \makecell{A positive integer, expressed in minutes}
            \midline
            \textbf{Ingredients} & A vector of $36$ elements representing the classes of ingredients. 
            The element in position $i$ is equal to $0$ if the corresponding ingredient is not used in the recipe; otherwise, it specifies how important the ingredient is in the recipe.
            \midline
            \textbf{Preparations} & \makecell{A vector of $9$ elements. Encoding is similar to the ingredients vector.}
            \midline
            \textbf{Link} & \makecell{A link pointing to the recipe on GialloZafferano}
            \botline
        \end{tabularx}}
        }
    \label{table:recipesDatasetFeatures}
\end{table}

{The resulting Recipes Dataset thus provides a mid-sized but carefully curated testbed: the number of recipes is large enough to support meaningful clustering, PCA, and neural-network training, while the ingredient taxonomy and manually assigned importance scores remain compact and interpretable. This is precisely the regime in which ILASP can be applied without prohibitive runtimes, and where the learned weak constraints can be understood in terms of intuitive culinary categories (e.g., ``pasta'', ``pork meat'', specific preparation methods).}

\subsection{Users Preferences Dataset}
A series of data regarding 54 user preferences over the recipes dataset were collected by means of a survey.
The most important criteria followed for the realization of this survey was the best trade-off between number of recipes and representativeness of the full distribution of the full set of 100 recipes.
To achieve the best results, we created 10 different surveys aiming to be as representative as possible of the distribution of the full set of 100 recipes, ensuring that all recipes appeared in at least some of the surveys.
To do so we used PCA for dimensionality reduction following \textit{kaiser rule} (taking into account components with eigenvalue $\ge$ 1) reducing the 47 features to 17 components.
Subsequently, we used the k-means clustering to partition the dataset, considering k = 2, 3, 4, 5, 20. 
k was finally set to 3 as it allowed to maximize the number of clusters, while minimizing the inter-group variance, excluding results in which we have group with small cardinality. 
From each of the three partitions, we draw randomly 7 recipes, which are combined to create a representative subset of 21 recipes. 
The 10 representative subsets obtained at the end of this process cover the entire recipes' dataset. 
Recipes can be present in multiple representative subsets, but not more than once in the same one. 
Each representative subset is used to create a survey, which is described in Table \ref{table:survey}.

\begin{table}[h!]
    \centering
    \caption{Survey structure}
    \resizebox{0.9\textwidth}{!}{
        {\tablefont\begin{tabularx}{\textwidth}{@{\extracolsep{\fill}} >{\centering\arraybackslash}m{0.25\textwidth} >{\centering\arraybackslash}m{0.7\textwidth}}
            \topline
            \textbf{Section} & \textbf{Description} \\ 
            \topline
            \textbf{Personal data} & \makecell{gender, age range and Italian region of residence}
            \midline
            \textbf{Recipes evaluation} & \makecell{each recipe presented is rated on a scale from 1 to 10, where \\1 means ``I would never eat it'' and 10 ``I like it very much''.}
            \midline
            \textbf{Sorting recipes} & \makecell{where the user has to sort the recipes presented according to \\her/his personal preferences.}
            \midline
            \textbf{Ingredient classes evaluation} & \makecell{for each ingredient class $c$, we ask to rate from 1 to 10 the\\ following question: ``Do you like recipes where there is $c$?''\\ where 1 means ``absolutely not'' and 10 ``absolutely yes''.}
            \midline
            \textbf{Ingredient meta-classes evaluation} & \makecell{for each ingredient meta-class $mc$, we ask to rate from 1 to 10\\ the following question: ``Do you like recipes where there is $mc$?''\\ where 1 means ``absolutely not'' and 10 ``absolutely yes''.}
            \midline
            \textbf{Preparations evaluation} & \makecell{for each preparations $p$, we ask to rate from 1 to 10 the\\ following question: ``Do you like recipes which require $p$?''\\ where 1 means ``absolutely not'' and 10 ``absolutely yes''.}
            \midline
            \textbf{Difficulty, preparation time and cost} & \makecell{we ask to rate from 1 to 10, where where 1 means\\ ``absolutely not'' and 10 ``absolutely yes'', the following\\ questions: ``Do you prefer simple cuisine to elaborate\\ cuisine?'', ``Are you willing to wait for a dish with a long\\ preparation time?'', and ``Are you willing to spend for a\\ good dish?''.}
            \midline
            \textbf{Ingredient/Preparation combination preferences} & \makecell{we ask the user if there are particular combinations of\\ ingredients (class or meta-class) and preparations for\\ which his preferences diverges from the scores assigned\\ to the individual elements (e.g.: the user has given a high\\ score to the fish ingredient meta-class and ``frying''\\ preparation method, but does not like fried fish). Each \\combination generated by the user also needs to be rated\\ (as in Recipes evaluation). We limited each combination\\ to a maximum of 3 items between ingredients and\\ preparations, and we collected for each user a maximum\\ of 4 combinations.}
            \botline
        \end{tabularx}}
        }
    \label{table:survey}
\end{table}

{The survey was administered to $54$ participants, each of whom completed exactly one of the ten questionnaires. This design yields a moderate number of users but a relatively dense preference signal for each user, which is adequate for our goal of studying ILASP-based explanations of per-user neural models rather than building a large-scale industrial recommender system.}

It is important to note that the section \textit{Sorting recipes} is divided in 3 subsections. Specifically, users are asked to sort 21 recipes in the survey, by effectively sorting 10 elements at a time. This allows on one side to avoid asking users to order too many items at once, on the other to introduce the uncertainty class described in Section \ref{subsection:Pref_Learn}.
These subsections are organized to maximize the number of common recipes among the ordering.
This would give us as much information as possible about the preferences and uncertainties of the user.
In fact, elements present in more than one subsection allow us to combine the three orderings ranked by the user, and create pairwise comparison between the 21 recipes, finding inconsistencies in the middle.
This process, which is deeply described in~\citep{fossemò2022inductive}, allow us to obtain uncertain pairs, and so passing from the classic pairwise binary classification problem, to the ternary one.
It's worth noting that this process would have been impossible to perform if we had used only one sort order, preventing us from inferring user uncertainties even if they are actually present.

Starting from the \textit{sorting recipes} section, we created the user preferences dataset.
For each entry (and so, for each user) there are 210 pairs, in which the elements are the ID of the recipes (pairs with elements with same ID were discarded).
Each pair has label 1 if the first element is preferred over the second, -1 if the second is preferred over the first, 0 in case of uncertain relationship among the two element.
The remaining sections were instead used as ground-truth in the experiment, as fully described in next chapters.
{In the released dataset, all responses are stored in anonymised form, and the mapping from raw survey answers to ternary pairwise labels is documented, so that other researchers can directly reuse the $11\,340$ labelled pairs or reconstruct alternative target variables (e.g., from ratings or meta-class evaluations) for related studies on interpretable preference learning and recommendation.}

\label{subsection:survey}

\section{Methods}
\label{subsection:methods}

\begin{figure}
    \centering
    \includegraphics[width=\linewidth]{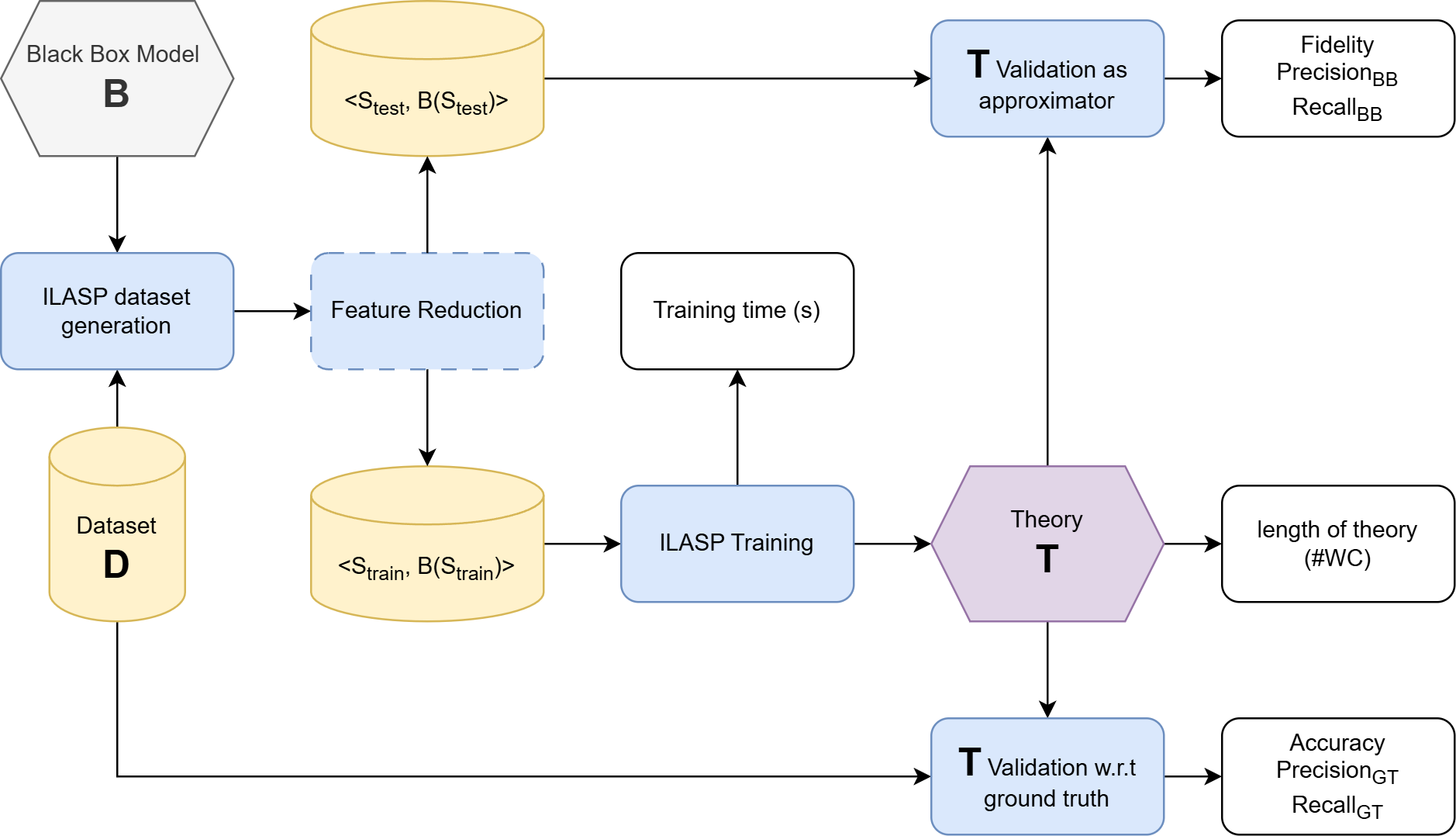}
    \vspace*{1mm}
    \caption{{Conceptual diagram of the overall approach, instantiating the approximation framework of Section~\ref{section:background}. Given a dataset $D$ and a pre-trained black box $B:X\to Y$ (in our case, a neural network), we first sample a training set $S_{\text{train}} \subseteq X$ and a test set $S_{\text{test}} \subseteq X$ uniformly at random from the feature space. These instances are then labelled with the corresponding black-box predictions $B(S_{\text{train}})$ and $B(S_{\text{test}})$, respectively. Dataset generation can follow either a \emph{global} strategy (Figure~\ref{subfig:global-approach}), where $S_{\text{train}}$ and $S_{\text{test}}$ are drawn from a task-relevant region $X_{\mathrm{rel}} \subseteq X$, or a \emph{local} strategy (Figure~\ref{subfig:local-approach}), where $S_{\text{train}}$ is built by perturbing a neighbourhood around specific query instances (cf.\ Definition of local explanation in Section~\ref{section:background}). On these labelled data we optionally apply feature reduction via direct or indirect PCA (Section~\ref{subsection:experimental-setup}) to control the size of the ILASP search space. ILASP is then run on $S_{\text{train}}$ with an appropriate language bias $L$, producing a symbolic theory $T$ consisting of weak constraints. At this stage we record the execution time and the length of the theory (i.e., the number of weak constraints in $T$) as proxies for computational cost and complexity $\Omega(T)$. Finally, $T$ is evaluated on $S_{\text{test}}$ to obtain Fidelity, Precision$_{BB}$ and Recall$_{BB}$ with respect to $B$, and on user ground truth (Section~\ref{subsection:gt-scores}) to obtain Accuracy\textsubscript{GT}, Precision\textsubscript{GT} and Recall\textsubscript{GT}, thus jointly assessing approximation quality and human alignment.}}
    \label{fig:approach-diagram}
\end{figure}

\begin{figure}[htbp]
    \centering
    \begin{subfigure}[c]{0.6\linewidth}
        \centering
        \includegraphics[width=\linewidth]{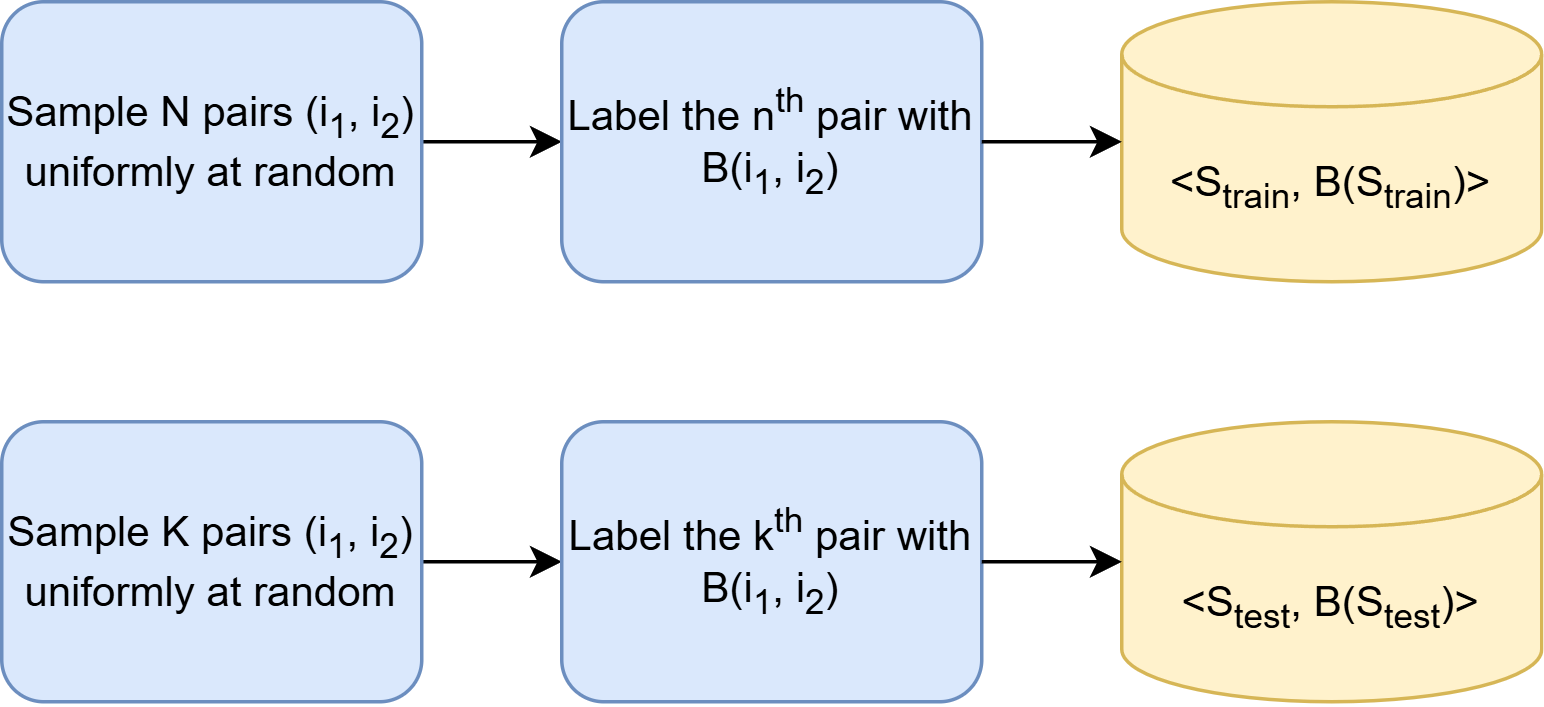}
        \caption{}
        \label{subfig:global-approach}
    \end{subfigure}
    \hfill
    \begin{subfigure}[c]{\linewidth}
        \centering
        \vspace*{1mm}
        \includegraphics[width=\linewidth]{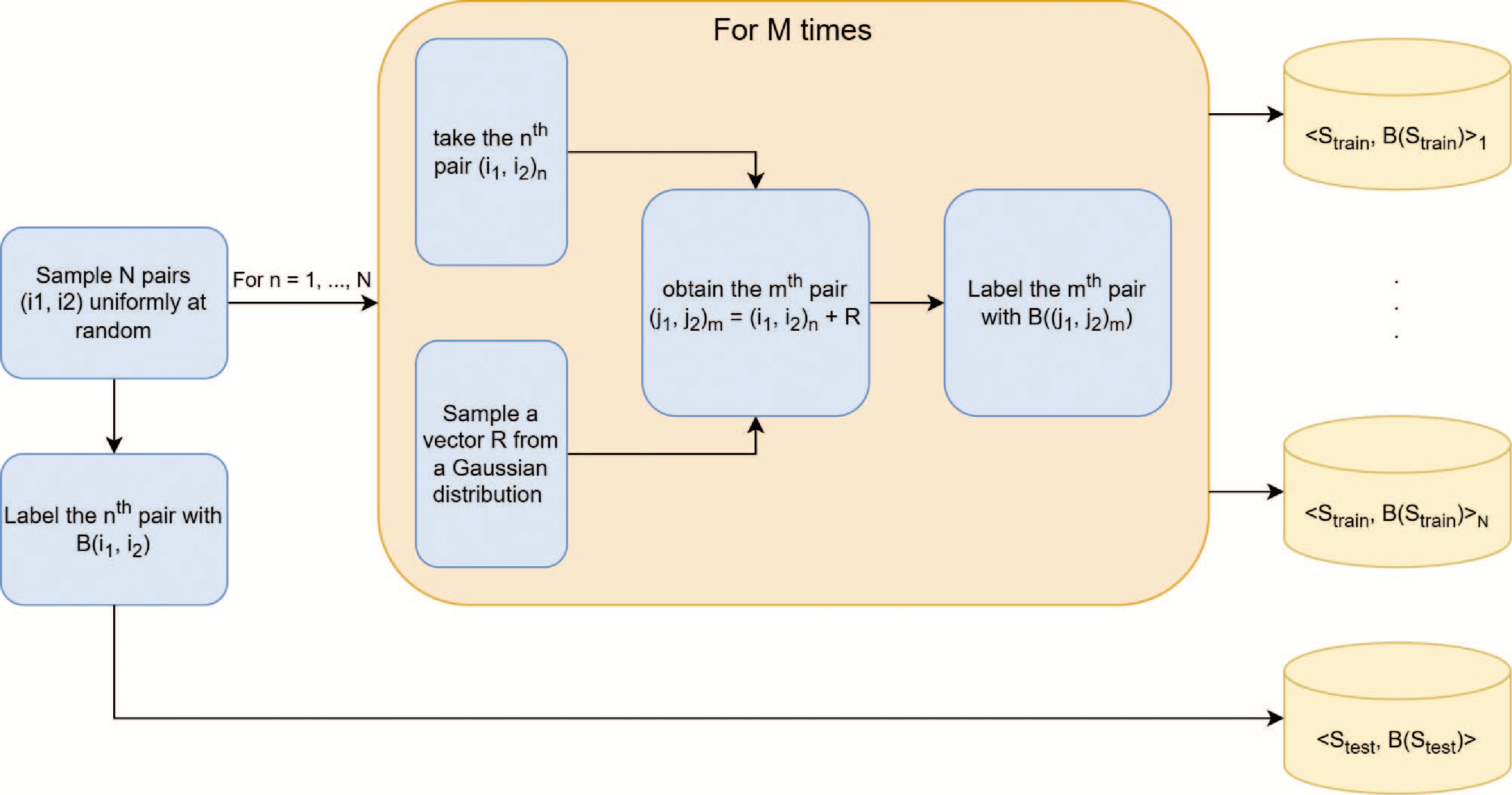}
        \caption{}
        \label{subfig:local-approach}
    \end{subfigure}
    \vspace*{1mm}
    \caption{{Generation of sampled datasets in the global (Figure~\ref{subfig:global-approach}) and local (Figure~\ref{subfig:local-approach}) approximation settings. In the \emph{global} case, we sample $N$ pairs $(i_1, i_2)$ uniformly at random from the feature space $X$ of ordered recipe pairs (cf.\ Section~\ref{subsection:Pref_Learn}) to form $S_{\text{train}}$, and we compute the corresponding labels $B(i_1,i_2)$ from the neural network $B$, obtaining $B(S_{\text{train}})$. A disjoint test set $S_{\text{test}}$ of size $K$ is generated analogously. This realises a global explanation in the sense of Section~\ref{section:background}, with region $A = X_{\mathrm{rel}} \subseteq X$. In the \emph{local} case, we first sample $N$ query pairs $Q_n = (i_1,i_2)_n$ that we wish to explain. For each query $Q_n$, we generate a local training set $S_{\text{train}}^{(n)}$ of $M$ perturbed pairs $(j_1,j_2)_m = (i_1,i_2)_n + R_m$ by adding Gaussian noise $R_m$ to the feature representation of $Q_n$ (Section~\ref{subsection:localmethod}). Each perturbed pair is labelled with $B(j_1,j_2)_m$, and the original query $Q_n$ together with its prediction $B(Q_n)$ is used as the corresponding test instance. This implements local explanations on metric balls $B_r(Q_n)$ around each query, as in the formal definition of local approximation.}}
    \label{fig:global-local-approach}
\end{figure}

In this section, we introduce methods used to approximate the underlying theory of a Neural Network~\citep{fossemò2022inductive} using ILASP with optimal efficiency.
The procedure is summarized in the Figure \ref{fig:approach-diagram}.
{Conceptually, our method instantiates the abstract approximation framework of Section~\ref{section:background}: the neural network $B$ plays the role of the black-box function $f$, ILASP theories $T\in\mathcal{T}$ are candidate white-box models, and our sampling procedures determine the regions $A\subseteq X$ on which $T$ is required to approximate $B$ (globally or locally).}
{Once the feature space $X$ and the logical vocabulary (predicates, types, and constants) are fixed for a given domain, the overall pipeline—sampling, labelling with $B$, ILASP training, and evaluation—depends only on access to the trained model $B$ and to unlabeled inputs in $X$. We do not require access to the original training set of $B$, which can be beneficial in scenarios where training data are sensitive or unavailable. However, ILASP’s background knowledge and language bias are necessarily instantiated from domain-specific information (e.g., the recipe attributes in Section~\ref{section:dataset}); in this sense, the ILASP components are \emph{tailored to the domain}, even though the methodological steps for approximation are independent of any particular dataset. When no real data are available but the structure of $X$ is known, one can still generate or collect a surrogate dataset of inputs~\citep{Angwin2016Compas} to query $B$ and drive ILASP.}

For the sake of completeness the neural Network architecture and training is reported in \ref{AppendixB}.
We tested two main approaches: \textit{global approximation} and \textit{local approximation}. Given that ILASP is not fast for highly dimensional datasets, we explored various methods to balance performance, execution time, and complexity of the resulting theories.

Our approach involved optimizing \texttt{\#maxp} ILASP hyperparameter to enhance efficiency. Additionally, we integrated Principal Component Analysis (PCA) to reduce dimensionality and further improve performance. 
To evaluate the degree of generalization of our explanation, we defined three bespoke scores based on Table \ref{table:survey} data: Accuracy\textsubscript{GT}, precision\textsubscript{GT} and recall\textsubscript{GT}.
These scores assess how well the theories align with the user's preferences obtained through a survey.

\subsection{Experimental setup}
\label{subsection:experimental-setup}

As we already mentioned, the main objective of this work is to approximate a Neural network (and, more in general, a black-box model) both globally and locally.
Recall that we deal with a user-dependent scenario, i.e., when we talk about approximating a Neural Network, we are talking about a Neural Network trained on the data from a specific user.
{Formally, for each user $\mathbf{x}$ we train a neural network $B_\mathbf{x}:X\to Y$ that approximates the pairwise preference function $p_\mathbf{x}(\lambda_i,\lambda_j)$ introduced in Section~\ref{subsection:Pref_Learn}. In our experiments, $X$ is the feature space of ordered recipe pairs (Section~\ref{section:dataset}), and $Y=\{-1,0,1\}$ encodes whether the first recipe is less preferred, equally preferred, or more preferred than the second.}

{Concretely, $B_\mathbf{x}$ is instantiated as a feed-forward Deep Neural Network with two dense hidden layers of 64 nodes each. A dropout layer with rate $0.1$ is applied after every dense layer, and batch normalisation is included between the two hidden layers. The network is trained with stochastic gradient descent using a learning rate of $0.0005$. We use a $\tanh$ activation at the input layer, ReLU in the first hidden layer, a linear activation in the second hidden layer, and a softmax activation in the output layer to model the ternary distribution over $\{-1,0,1\}$. Additional architectural and training details (e.g.\ early stopping, class balancing, and train/validation splits) are reported in \ref{AppendixB}. On held-out data, these models reach on average 82.72\% accuracy, 83.26\% precision and 82.43\% recall across users, which we take as a reasonable performance baseline for the subsequent approximation by ILASP.}

{In the remainder of this section, we describe the \emph{general} global and local approximation procedures, deliberately abstracting away from specific training-set sizes $N$, $K$ and $M$ or from particular PCA configurations. The concrete hyperparameter choices for our experiments are reported in Section~\ref{subsection:hyperparameters}.}

\subsubsection{Global Approximation}

In the global case, we are interested in extrapolating the general logic that the Neural Network uses to predict user preferences.
The procedure is summarized in Figures \ref{fig:approach-diagram} and \ref{subfig:global-approach}.
Specifically, we sample \textit{N} pairs \textit{($i_1$, $i_2$)} uniformly at random from the feature space, and label these pairs according to predictions \textit{B($i_1$, $i_2$)} of the Neural Network.
We repeated the same procedure to sample and label a test set of $K$ samples
Then, we used ILASP on the training set using an appropriate language bias \textit{L} in order to obtain an explanation theory. Finally, we test the learned theory on the test set.

{In terms of the approximation framework of Section~\ref{section:background}, global approximation corresponds to choosing a task-relevant region $X_{\mathrm{rel}}\subseteq X$ (here, the support of the recipe-pair distribution) and seeking an ILASP theory $T\in\mathcal{T}$ that approximately minimises $L(T,B,X_{\mathrm{rel}})$ while keeping the complexity $\Omega(T)$ (e.g.\ number of weak constraints) below a threshold. The training set $S_{\text{train}}$ provides a finite sample from $X_{\mathrm{rel}}$ on which ILASP can observe the behaviour of $B$, while $S_{\text{test}}$ is used to estimate fidelity, Precision$_{BB}$ and Recall$_{BB}$ as empirical proxies for $L(T,B,X_{\mathrm{rel}})$.}

{At the ILASP level, the background knowledge $B$ encodes the feature representation of each recipe (Section~\ref{section:dataset}), while the language bias $L$ specifies admissible weak-constraint patterns over these features (cf.\ Section~\ref{par:ILASP}). Brave ordering examples encode the pairwise preferences predicted by the neural network: for each sampled pair $(i_1,i_2)$, an ordering example asserts that the answer sets extending $i_1$ should be preferred, tied, or less preferred than those extending $i_2$, depending on whether $B(i_1,i_2)=1,0,-1$ (with the ILASP-specific convention between $>$, $<$ and $\sim$, see equation~\eqref{eq:Tclass}). The concrete ILASP encoding for the global case, including mode declarations and ordering examples, is reported in \ref{AppendixD}.}

{A typical theory learned by ILASP in the global setting may look as follows.}

\begin{example}
\leavevmode\vspace{0ex}
\begin{verbatim}
    :~ value(vegetables, V1).[-V1@1, V1]
    :~ value(meat, V1).[-V1@2, V1]
    :~ value(difficulty, V1).[-V1@3, V1]
    :~ value(stewing, V1).[V1@4, V1]
    :~ value(dairies, V1), category(3).[V1@5, V1]
\end{verbatim}

\end{example}

{Abstractly, this theory consists of weak constraints whose bodies mention simple conjunctions of feature atoms (such as \texttt{value(vegetables,V1)} or \texttt{value(dairies,V1), category(3)}) and whose weights and priority levels encode how these features contribute to the preference relation $\succ_\mathbf{x}$ defined in equation~\eqref{eq:Tclass}. Negative weights (\texttt{-V1}) reward higher values of the corresponding feature at a given priority level, while positive weights (\texttt{V1}) penalize them. In this example, the presence of vegetables and meat, as well as higher difficulty, tends to increase preference (at different priority levels), whereas the use of stewing and dairies in first-course recipes (category 3) is penalized.}

{From the perspective of preference learning (Section~\ref{subsection:Pref_Learn}), the learned theory $T$ induces an ordering over recipes by comparing the costs of their associated answer sets. When restricted to pairwise comparisons $T(\lambda_i,\lambda_j)$ as in equation~\eqref{eq:Tclass}, this yields a symbolic surrogate of the neural network’s pairwise predictions. As discussed at the end of Section~\ref{par:ILASP}, the weak constraints in such theories can be systematically translated into natural-language statements, providing human-understandable explanations of the ranking logic captured by the black-box.}

\subsubsection{Local Approximation}

In the local case, we are interested in the reason that led the Neural Network to make a specific predicition for a query \textit{Q}. 
The procedure is summarized in Figures \ref{fig:approach-diagram} and \ref{subfig:local-approach}
As before, we sample \textit{N} pairs $(i_1, i_2)$ uniformly at random from the feature space, and we label them with the Neural Network predictions $B(i_1, i_2)$. For each generated pair $Q$, we also generate \textit{M} associated pairs $(j_1, j_2)_m$ \textit{sampled around} $Q$, and label them with Neural Network predictions $B(j_1, j_2)_m$. We chose to implement \textit{sampling around} a pair $(j_1, j_2)_m$ as adding Gaussian noise to $i_1$ and $i_2$, but one can also use other noise models. {Note that depending on the chosen noise model and feature type (categorical vs continuous), some post-processing may be necessary to project perturbed points back into the valid feature space, as discussed in Section~\ref{subsection:results}.}

{From the viewpoint of Section~\ref{section:background}, local approximation corresponds to choosing, for each query $Q$, a metric $d$ on $X$ and a radius $r>0$ such that the perturbations $(j_1,j_2)_m$ lie in the ball $B_r(Q) = \{x\in X \mid d(x,Q)\le r\}$. ILASP is then asked to find a low-complexity theory $T_Q$ that approximates $B$ \emph{restricted to} this local region, i.e., that nearly minimises $L(T_Q,B,B_r(Q))$ under the complexity constraint $\Omega(T_Q)\le\kappa$. In our implementation, the metric $d$ is induced by $\pi_Q$ defined below, and the brave ordering penalties are modulated by $\pi_Q$ so that ILASP preferentially fits pairs closer to $Q$.}

Let $\pi_Q$ be a metric distance associated to the query \textit{Q}, then for any $(j_1, j_2)$ there is an associated $\pi_Q(j_1, j_2).$
Formally, we define \textit{$\pi_Q$($j_1$, $j_2$)} as follows:
\begin{equation}
    \pi_Q(j_1, j_2) = \sum_{k=1,2} \sqrt{\sum_{f}(d(j_{k}(f), i_{k}(f)))^{2}}
    \label{eq:metricDistance}
\end{equation}
where \textit{$j_k(f)$} is the value of feature \textit{f} for the item \textit{$j_k$}, and \textit{$d(j_{k}(f), i_{k}(f))$} evaluates to \textit{$j_k(f) - i_k(f)$} when \textit{f} is a continuous feature, whereas it evaluates to
\begin{equation}
    d(j_{k}(f), i_{k}(f))=\left\{
    \begin{array}{@{}ll@{}}
        0, & \text{if $j_k(f) = i_k(f)$} \\
        3, & \text{otherwise}
    \end{array}\right.
\end{equation}
when \textit{f} is a categorical feature.
We chose 3 as an arbitrary value for the distance between categorical features, as ILASP works with integer weights, and 3 was a reasonable choice given the ranges of the dataset's continuous features.
For each of the \textit{N} queries, we use ILASP on the corresponding generated \textit{M} pairs using an appropriate language bias \textit{L}, by making ILASP prioritize theories that cover the pairs which are closer to the query (i.e., by putting  \verb|o_n| = $\pi_Q(j_1, j_2)$ on the corresponding \verb|brave_ordering| of the pair ($j_1, j_2$), {as detailed in \ref{AppendixD}}).
Note that the original query is not included in the training set, but it is always included in the test set.
Thus, differently from the global approximation, the cardinality of the test set is $K = N$, and the test samples are exactly the $N$ pairs sampled at the start of the local sampling procedure. On the other hands, the training set size is identified by $M$, which is the number of pairs generated through Gaussian noise.

{In summary, the global and local variants differ only in how the region $A$ is instantiated (whole task region vs.\ neighbourhood of a query) and how training examples are weighted or selected within ILASP. The underlying ILASP formalism (background knowledge, language bias, and use of brave orderings to encode preferences) remains the same, which makes the method amenable to other pairwise preference domains beyond recipes, provided that a structured feature representation similar to that in Section~\ref{section:dataset} is available.}

\label{subsection:localmethod}
\subsection{Principal Component Analysis}

We use PCA as the main method to reduce execution time while maintaining high performance~\citep{Abdi2010PCA}. We make typical use of PCA to decrease the feature space size with little information loss. 
In fact, PCA is designed to store in the first Principal Components as much variance (of original data) as possible, and so to capturing as much information as possible. Considering the use of \textit{kaiser rule} for PC selection, this lead to a negligible loss of information.
Note that this is also reflected by the results in Section \ref{subsection:results}, where we get an overall increase in estimators when using PCA rather than not using at all.
We explored two different techniques, which we refer to as \textit{indirect} and \textit{direct}.

In what we call \textit{indirect PCA}, we determine the most important features among the first \textit{n} Principal Components (PCs), so that we can reduce the number of features in the dataset accordingly.
Let us consider the first \textit{n} principal components obtained by applying the PCA on the dataset, and the weights $w_{ij}$ that the feature \textit{i} assigns to component \textit{j}, then we only kept the features such that:
\begin{equation}
    |w_{ij}| \ge \mu'_{j} + 2\sigma'_j  \; \; \; \; \; \; \textit{j} = 1, ..., \textit{n}
    \label{eq:indirectlyPCA}
\end{equation}
Where $\mu'_{j}$ and $\sigma'_j$ are respectively the mean and the standard deviation of $|w_{j}|$.
This means that, the features whose absolute weights are greater than the mean of the features' absolute weights plus two times the standard deviation, among the first \textit{n} PCs, are those that we keep in the dataset.

\begin{figure}[]
    \includegraphics[width=\textwidth]{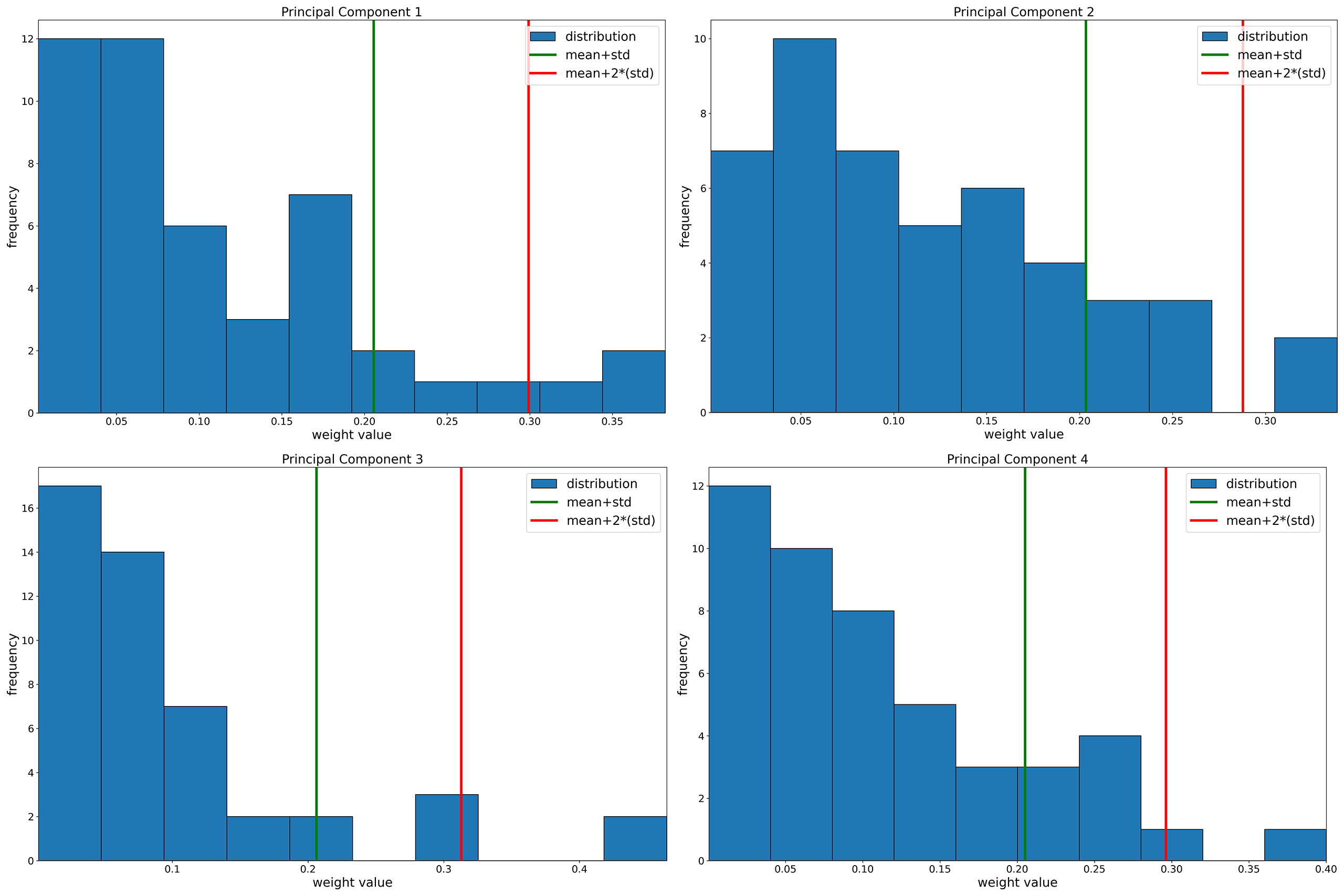}
    \caption{Histogram of the absolute weights of the features for the first four principal components. When using \textit{indirect PCA}, we select the features with weights in the $i^{th}$ PC that are to the right of the red line.}
    \label{fig:pcaIndirect}
\end{figure}

\noindent In Figure \ref{fig:pcaIndirect} we show the histograms of the features' absolute weights for the first four PCs. Distributions of the PCs fit well with a $\mathit{Gamma}(k,\theta)$ distribution with shape parameter $k\approx 1.3$ and scale parameter $\theta\approx 0.1$ ($p<0.05$ according to the Kolmogorov-Smirnov test for most PCs).
This indicates that \eqref{eq:indirectlyPCA} selects approximately $5\%$ of the features.

In the \textit{direct} approach, the dataset is passed on to ILASP directly after the PCA, i.e., we pass the PCs as features.
Differently from the \textit{indirect} approach, the direct approach does not exclude any feature since, by the very definition of PCA, they all contribute towards each of the PC. 
It follows that the effectiveness of this technique is influenced by how and how many features are retro-projected from the PC involved in the weak constraint of the obtained theories.
There are several ways to perform retro-projection; {for instance, to remain consistent with the \textit{indirect} approach and other experiments, we extrapolate features satisfying \eqref{eq:indirectlyPCA} with $n = 8$, i.e.\ we first select the principal components whose eigenvalues are at least $1$ according to the Kaiser rule (the same criterion used in Section~\ref{section:dataset}), and then, for each of these $8$ PCs, we retain the original features whose weights are more than two standard deviations above the mean in absolute value. This threshold reflects the empirical Gamma-like distribution of weights in Figure~\ref{fig:pcaIndirect} and typically results in a sparse subset of highly influential features per PC.}
{In other words, indirect PCA drops low-loading features \emph{before} ILASP training, shrinking the search space $S_L$ and thus the optimisation problem; direct PCA compresses the feature space into PCs for ILASP training and then uses retro-projection only at the \emph{explanation} stage to map PCs back to a small set of original interpretable features. In both cases, the goal is to keep the resulting ILASP theory $T$ both computationally tractable (short runtimes) and interpretable (small $\Omega(T)$ expressed in terms of domain-level attributes such as ingredient meta-classes and preparations).}

It is worth noting that, while in the \textit{indirect} approach we drop a subset of the original features before using ILASP, the retro-projection excludes features after ILASP application, i.e., when generating explanations. In fact, the explanation should include only the essential features needed to capture the variance in the data. This is necessary for two reasons: we want to keep the explanation concise, and we want to ensure that it is easily understandable to humans.

Finally, assuming that the trained black-box model is on hand and that, in the case of \textit{direct} or \textit{indirect} PCA, the corresponding mixing matrix or feature-selection mask has been retained, one can reply the overall proposed approach. 
In fact, one can generate arbitrary synthetic samples, apply either \textit{direct} or \textit{indirect} PCA, and query model for predictions.
In turn, these data can be then used to generate a theory using ILASP, following the methodology described in Section \ref{subsection:experimental-setup}.

\subsection{User ground-truth scores}
Symbolic artificial intelligence, such as ILASP, has a well-known advantage over black-box models. It can reason over the theory it generates, rather than opaquely making predictions on new data points.
We use the transparent output theory to better evaluate ILASP performance.

{As discussed in Section~\ref{section:background} (see in particular equation~\eqref{eq:Tclass} and the subsequent discussion on fidelity~\citep{gui18survey}), we use \emph{fidelity} to quantify how closely an interpretable model $T$ mimics the predictions of the black box $B$ on a given test region $A$. Operationally, fidelity is computed like an accuracy score, but with respect to $B$’s outputs rather than ground-truth labels. While this makes the notion similar in form to standard accuracy, the underlying semantics differ: accuracy measures predictive performance against the true preference function $p_\mathbf{x}$, whereas fidelity measures agreement with $B$, regardless of whether $B$ is itself accurate. Because there is no widely adopted naming convention for precision and recall in this context, we use Precision$_{BB}$ and Recall$_{BB}$ to denote precision and recall scores computed on ILASP’s predictions against the outputs of the black box.}

Furthermore, as described in Section \ref{subsection:experimental-setup}, each test set is sampled from the feature space.
For this reason, we do not have ground-truth labels of test sets, except for those provided by the black-box model.
This scenario is quite common in real-world use cases, where data may often be unlabeled.
Thus, we propose the use of ground-truth data described in Table \ref{table:survey} to define Accuracy\textsubscript{GT}, precision\textsubscript{GT}, and recall\textsubscript{GT}, that are score which help us to better understand the degree of generalization of our explanation.
In fact, Fidelity, precision\textsubscript{BB} and recall\textsubscript{BB} tell us only how well the explanation is approximating the BB (which remain the main task of the paper), while these new score tell us how much this theory is in line with the preferences expressed by the users in the survey in general.
We do this by comparing the labels assigned by the explanation returned by ILASP with those that would be assigned if ground-truth scores were used instead of the weights and the priority levels provided by ILASP in the explanation.
{It should be noted that there is no single canonical way to define how labels would be derived from ground truth in such a symbolic setting; the procedure we propose below is one principled way of mapping user scores to weak-constraint priorities so as to obtain a reference theory $T_{GT}$ for comparison. Alternative mappings would be possible if different assumptions about priority scales or score ranges were adopted.}
Recall from Table \ref{table:survey} that recipe features are ranked by users in the integer range $[1,10]$.
For a user $x$ and a feature $f$, we denote the ground-truth evaluation of $f$ according to $x$ by $G_x(f)$.
Then, given a theory $T$ returned by ILASP after its application, which is composed of only weak constraints, for each feature $f \in T$ we define a weak constraint $wc_{GT}(f)$ as:
\\

\verb|:~ value(|$f$\verb|, V1)[V1@|$|m(\bar{G}_{x}(f))|$\verb|, V1].|\;\;\;\;\;\;\;\;\;\;\: if $m(\bar{G}_{x}(f)) > 0$

\verb|:~ value(|$f$\verb|, V1)[-V1@|$|m(\bar{G}_{x}(f))|$\verb|, V1].|\;\;\;\;\;\;\;\;\; if $m(\bar{G}_{x}(f)) < 0$ \\

\noindent where:

\begin{equation}
    \bar{G}_{x}(f) = (G_{x}(f) - 5)(-1)
\end{equation}

\begin{equation}
    m(\bar{G}_{x}(f)) = 
    \begin{cases}
        \bar{G}_{x}(f) \;\;\;\;\;\;\;\;\;\, if \;\: \bar{G}_{x}(f) > 0\\
        \bar{G}_{x}(f) -1 \;\;\; otherwise\\
    \end{cases}
\end{equation}

\noindent the mapping function $m(\bar{G}_x(f))$ for $G_x(f)$ is used to align with ILASP’s weak constraint semantics.
Specifically, $G_x(f)$ belongs to the range [1,10], where a value of 1 means that the user $x$ absolutely dislikes $f$, while a value of 10 indicates a strong preference for $f$.
Such score is used to define weight sign and priority level of the new weak constraints, assigning progressively higher priorities to the weak constraint if $m(\bar{G}_{x}(f)) > 0$ and vice versa, with the rationale that the score given by the user implicitly define a priority level among features.
To define the priority level of the weak constraint, we use the mapping function $m(\cdot)$.
Since $G_x(f) \in [1, 10]$ and we need a priority level in [1, 5](since \texttt{\#maxp} = 5), discerning between positive and negative preferences (i.e. negative and positive weight sign with regard to Section \ref{subsection:ASP}) we define $m(\cdot)$ on $\bar{G}_x(f)$.
$\bar{G}_x(f)$ is obtained centering $G_x(f)$ around zero and inverting the sign, so that $\bar{G}_x(f) \ge 6$ results in negative weights in weak constraint, and vice versa.
In our case, working with \texttt{\#maxp} = 5, we simply set $m(\bar{G}_x(f) = \bar{G}_x(f))$ if $\bar{G}_x(f) > 0$.
However, since it wouldn’t make sense to assign a priority level of zero when $\bar{G}_x(f) = 0$ (thus $G_x(f) = 5$), we shift by one whenever $\bar{G}_x(f) \le 0$.
Note that the resulting priority is set in absolute value to avoid negative priority levels. 
It’s important to note that the choice of $m(\cdot)$ is not unique and can be adapted depending on the context (for example, with a different value of \texttt{\#maxp}).

Finally, given a theory T, we define $T_{GT} = GT(T)$, where:

\begin{equation}
    GT(T) = \{wc_{GT}(f)|f \in T\}
\end{equation}

\noindent So, accuracy\textsubscript{GT}, precision\textsubscript{GT} and recall\textsubscript{GT} are calculated comparing the label assigned by the explanation $T$ given by ILASP after its application with the label assigned by $T_{GT} = GT(T)$, following the cost concept described in Section \ref{subsection:ASP}. 

To better understand the process, consider the following example:
Let $T$ = \{\verb|:~value(browning,V1)[V1@1, V1].|\} (where \textit{browning} is a preparation of a recipes, as described in Table \ref{table:recipesDatasetFeatures}) be {the theory returned by ILASP}, then $\mathcal{F} = \{f|f \in T\} = \{browning\}$.
Let be $G_x(browning) = 3$, then $\bar{G}_x(browning) = (3 - 5)(-1) = 2$, and so {$m(\bar{G}_x(browning)) = 2$}.
So, {$T_{GT}$ = $GT(T)$ = }\{\verb|:~value(browning,V1)[V1@2, V1].|\}.
{Note that while in $T$ the feature \textit{browning} is placed at the first level of priority, in $T_{GT}$ it appears at the second level. As we will show later in this example, this does not necessarily mean that $T$ and $T_{GT}$ will produce different results.}
Consider the sample, $s = (r_1, r_2)$ which is a pair of recipes we intend to classify with regard to user $x$ preferences.
Since our theories are composed only of the feature ``browning'', let's focus only on that feature.
Let's say that $r_1(browning) = 3$, while $r_2(browning) = 1$ and consider the theory $T$, then we get the cost vectors $p_{T}(r_{1}) = [0\;\,3]$ and $p_{T}(r_{2}) = [0\;\,1]$.
So, with regard to the theory $T$, the user $x$ prefers $r_2$ over $r_1$ (and so assign the label 1).
Consider now the theory $T_{GT}$, then we get {$p_{T_{GT}}(r_{1}) = [3\;\,0]$ and $p_{T}(r_{2}) = [1\;\,0]$}, which return the same label, confirming that theory $T$ weight and priority level assignments are in line with the ground-truth available data.

Consider the previous example, but now suppose that $G_x(browning) = 7$, then $\bar{G}_x(browning) = (7 - 5)(-1) = -2$ {and so $m(\bar{G}_x(browning)) = -3$. Finally we obtain that $GT(T)$ = }\{\verb|:~value(browning,-V1)[V1@2, V1].|\}.
In this case {$p_{T_{GT}}(r_{1}) = [-3\;\,0]$ and $p_{T}(r_{2}) = [-1\;\,0]$}, and so returned label would be -1, disconfirming that theory $T$ weight and priority level assignments are in line with the ground-truth available data.

\label{subsection:gt-scores}

\section{Results}
In this section, we {answer to the research question made in Section \ref{section:introduction}, and} present the results of our proposed methods.
Out of all the users, we focused on $10$ of them whose corresponding neural networks had the best accuracies. The results involve the use of ILASP both as a global and local approximator and the PCA as a \textit{direct} and \textit{indirect} method.

To further decrease dimensionality, we considered different ingredient granularities, i.e., standard classes for neural networks and meta-classes for ILASP as illustrated in Table \ref{table:ingredients-classes}.
The results report on average fidelity, precision\textsubscript{BB}, recall\textsubscript{BB}, and execution time for each case, as well as the average number of weak constraints returned by ILASP after its application and the average accuracy\textsubscript{GT}, precision\textsubscript{GT}, recall\textsubscript{GT}.

The dataset is unbalanced, with an average of $17.44\%$ of pairs labeled as \textit{uncertain} for each user. This affects the neural network's performance, reducing precision and recall, while accuracy is less impacted as it accounts for both true positives and true negatives.

{Experiments have been made on machine with AMD Ryzer 9 5900X CPU, 32GB RAM, GeForce RTX 3080 GPU.}
\paragraph{Global approximation.} {In the global approximation approach we considered training sets of different sizes, namely with $N = 45$, $105$ and $190$ pairs; on the other hand, the test set (which is the same for all three training sets) has $K = 105$ pairs.}
\paragraph{Local approximation.} {In the local approximation approach, we uniformly sampled a set of $N = 100$ pairs and then generated a training set for each pair in this set, i.e. $M = 45$, $105$ and $190$.
The results are calculated as average performance on pairs for each user.
As described in \ref{subsection:localmethod}, the generated training set are obtained by adding gaussian noise. To examine how ILASP behaves under various degrees of locality, we carried out experiments using different standard deviation values for the Gaussian noise. Specifically, we tested for \textit{$\sigma$} values of $1$, $0.1$, $0.01$, and $0.001$.
}

{It must be noted that in our dataset the recipe features are non-negative float numbers (see Table \ref{table:recipesDatasetFeatures}). However, since \texttt{clingo} only supports integers, we must preprocess data. We first shift the range of the values so that $0$ becomes the minimum. This keeps data consistent with the initial dataset, as Gaussian noise can introduce negative values. Next, we generate the Gaussian noise with a standard deviation equal to $0.001$, $0.01$, $0.1$, or $1$, as displayed in Tables \ref{table:resultsLocalData1}, \ref{table:resultsLocalData2}, and \ref{table:resultsLocalData3} in Section \ref{subsection:local-results}.
To reason over integers as required by \texttt{clingo}, we must round up the obtained values. However, due to the magnitude of noise (for standard deviation equal to $0.001$, $0.01$, and $0.1$), rounding up typically results in discarding the changes made. For this reason, we multiplied the values by a factor inversely proportional to the standard deviation, so that the system is sensitive to all variations. For instance, for standard deviation $0.1$, we multiply by $10$; for standard deviation $0.01$, we multiply by $100$, and so forth.}
\paragraph{Indirect PCA.} In the \textit{indirect} approach, we experimented with \textit{n} equal to $8$ and $17$. {We chose two different \textit{n} to conduct more complete study, comparing \textit{indirect} approach with different value of \textit{n}}. We {specifically} chose $17$ as it is the number of Principal Components involved during the training of neural networks, and $8$ as it is its rounded half. Note that $17$ PCs explain about $76\%$ of the dataset and preserve $34$ features, while $8$ PCs explain about $49\%$ of the dataset and preserve $15$ out of $48$ original features. 
\paragraph{Direct PCA.} In the \textit{direct} approach, we tested the case with $5$, $10$, $15$ and $20$ PCs.
Finally, following discussion from Section \ref{subsection:hyperparameters}, \texttt{\#maxp} was set to $5$.

\label{subsection:results}

\subsection{Preliminary analysis}\label{subsection:hyperparameters}
To benchmark ILASP performance, we conducted a study to understand the role of ILASP hyperparameters in terms of \textit{accuracy}, \textit{precision}, \textit{recall} and \textit{execution time}. Since these experiments usually require a long execution time, we focused only on \texttt{\#maxp} (maximum number of weak constraints in a theory), setting \texttt{\#maxv = 1} (maximum number of variables for weak constraints in a theory). To further fix environmental conditions, we ran experiments in the hard case of global approximation, indirect PCA, and $n = 8$. 

\begin{figure}[]
	\centering
	\includegraphics[width=\textwidth]{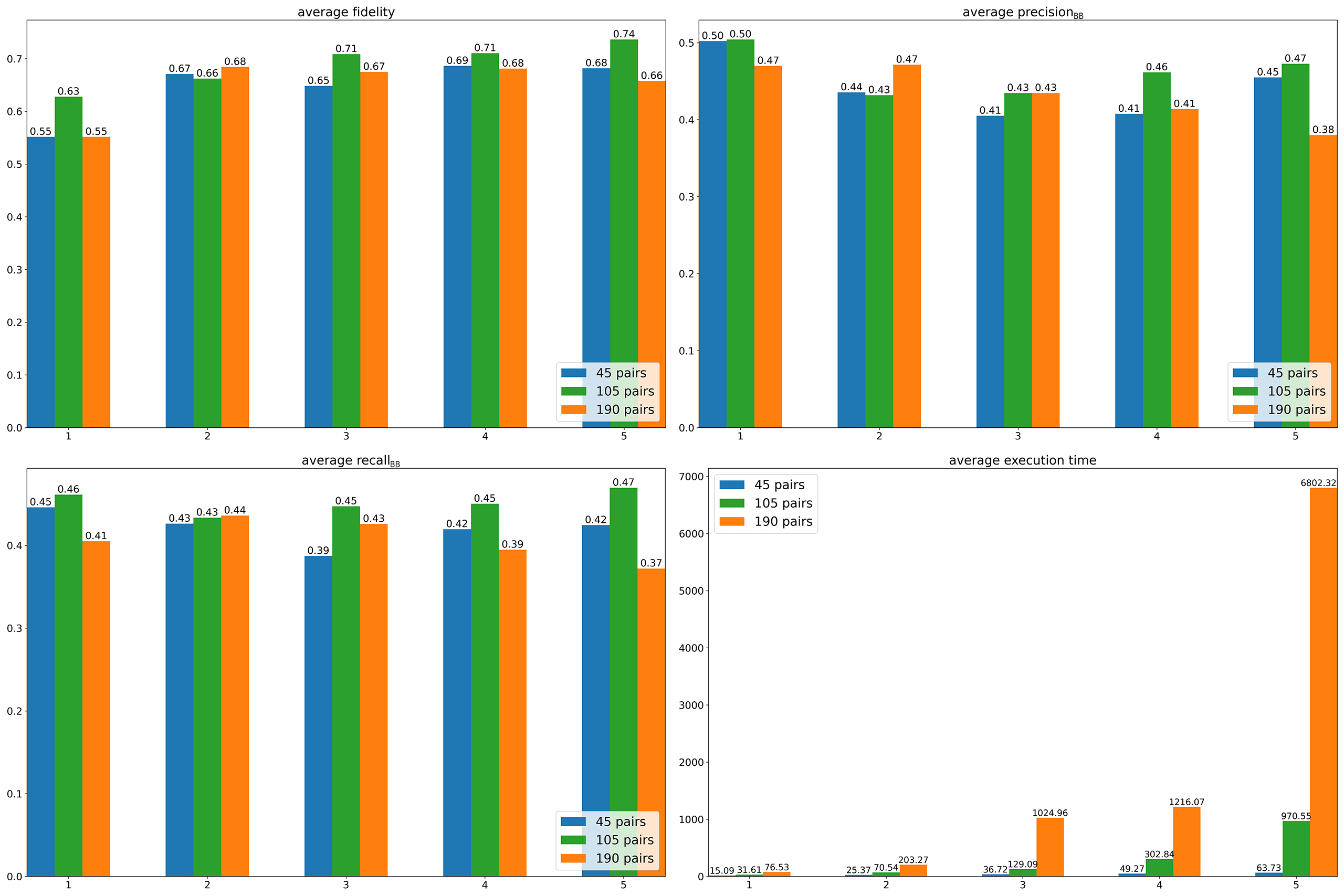}
	\caption{Global fidelity, precision\textsubscript{BB}, recall\textsubscript{BB}, and execution time results for \texttt{\#maxp} between $1$ and $5$, \textit{indirect} PCA}
	\label{fig:all_parameters8PC2STD}
\end{figure}

As it can be seen in Figures \ref{fig:all_parameters8PC2STD}, the execution time grows exponentially both in the size of the dataset and \texttt{\#maxp}.
As can be seen from these figures, \texttt{\#maxp} also affects fidelity, precision\textsubscript{BB}, and recall\textsubscript{BB}. With the increase of \texttt{\#maxp}, fidelity rises by approximately 11 to 13\%, while precision\textsubscript{BB} and recall\textsubscript{BB} generally decline by around 2 to 9\%.
{However, overall, we did not notice a significant trend in the obsserved metrics}.
It is worth noting that better results were generally obtained with the training set consisting of 105 pairs, compared to the one with 190 pairs. 
To understand the reasons behind these results, we analyzed the similarity between the sampled training sets used for ILASP and the dataset on which the black-box model was trained, using the Maximum Mean Discrepancy (MMD) metric~\citep{gretton2012MMD}. 
This analysis revealed that the training set with 105 pairs is more similar to the neural network's training data, whereas the set with 190 pairs is less similar.
This suggests that ILASP may perform less effectively when applied on the set with 190 pairs, possibly due to inconsistencies in the ground truth derived from the neural network, which could be more prone to errors when classifying unfamiliar pairs. Importantly, this outcome seems to be a coincidence, as the training sets were randomly sampled.
Considering the differences in execution time, particularly with the dataset containing 190 pairs, the best choice for the experiments would be to set \texttt{\#maxp} = 4, as it offers the optimal balance between execution time and performance.
However, in the remainder of this paper we set $\texttt{\#maxp} = 5$, as we were interested in complex theories; nonetheless, other choices for $\texttt{\#maxp}$ are possible, and Figures \ref{fig:all_parameters8PC2STD} serves as a useful guide to help the reader select the most appropriate value depending on the specific application.

\subsection{Global approximation}
\begin{figure}
    \centering
    \includegraphics[width=\linewidth]{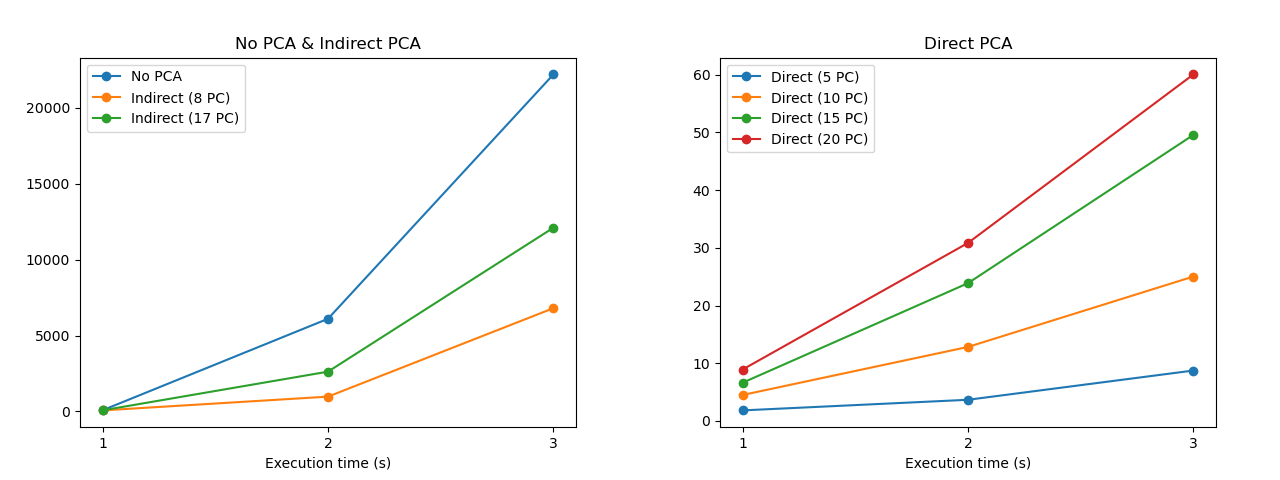}
    \vspace*{1mm}
    \caption{{Execution time comparison for global approximation across No PCA, Indirect PCA, and Direct PCA. Indirect PCA preserves the exponential growth observed with No PCA, albeit at a reduced rate; Direct PCA instead exhibits an approximately linear scaling, with the slope decreasing as fewer principal components (PCs) are retained.}}
    \label{fig:direct&indirect_execution_time}
\end{figure}

\paragraph{General results.} The results of ILASP globally approximating the neural network on the three different training sets are reported in Tables \ref{table:resultsGlobalData1}, \ref{table:resultsGlobalData2} and \ref{table:resultsGlobalData3}.
{Tables show that it is possible to make the black-box model globally transparent, achieving a fidelity of approximately 70\%, thereby \textbf{Q1} and \textbf{Q2}.}
As expected, using PCA decreases the computational time. What is most interesting is the fact that
both \textit{direct} and \textit{indirect} PCA approaches lead to an overall increase in estimators compared to the scenario without PCA. 
However, in terms of user ground-truth scores, we generally achieve superior results only in \textit{indirect} PCA approach, while \textit{direct} remain stable only on accuracy\textsubscript{GT}.
\paragraph{Scalability of the method.} In terms of execution time, {as visible in Figure \ref{fig:direct&indirect_execution_time}}, there is a significant difference between the different approaches: \textit{indirect} PCA reduce the execution time but maintain the exponential growth between datasets (similarly to the case without PCA), while in the \textit{direct} PCA the time increase is more like to be linear.
Additionally, the \textit{direct} PCA approach generally provides better estimators. 
Although counterintuitive, the difference in results between the \textit{direct} approach and other cases cannot be attributed to a greater decrease in feature space, as \textit{indirect} PCA involves only 15 features (when \textit{n} = 8), whereas the \textit{direct} approach involves 20 features. 
This outcome results directly from PCA’s definition as a change of basis: it re-expresses the original feature space in a new orthonormal basis optimized to capture variance \cite{Abdi2010PCA}.
When ILASP is applied using the \textit{direct} PCA case, it has a simpler feature space compared to the starting one. 
This assumption does not hold in the case of \textit{indirect} PCA, where dimensionality reduction simply involves removing features that do not meet the criteria defined by (\ref{eq:indirectlyPCA}). 

{Overall, the reported results indicate that both the indirect and direct approaches successfully enhance the scalability of the method, thus answering \textbf{Q3}.}
\paragraph{Performances of PCA approaches.} In the \textit{direct} PCA approach, we generally observe a gradual improvement in fidelity, precision\textsubscript{BB}, and recall\textsubscript{BB} up to 10 principal components (PC), after which they decrease.
This could be a form of early overfitting caused by the fact that the combined use of the \textit{direct} PCA method and the predictions of the neural network as labels oversimplifies the feature space and ground-truth.
{Moreover, }the theory returned in \textit{direct} cases have less complexity, reaching the average max number of weak constraint of 1.9.
{As it is visible from the tables}, we obtain the worse ground-truth scores, with only accuracy\textsubscript{GT} being comparable to other cases.
This result could be attributed to a combination of the two earlier suppositions: on one hand, the simplified feature space encourages the development of streamlined theories (with fewer weak constraints), and on the other, the early overfitting leads to theories that are highly tailored to the training set, potentially limiting their ability to fully capture the ground-truth.

On the other hand, \textit{Indirect} PCA approach obtains more complex theory and better ground-truth scores, both generally better than the results obtained without the use of the PCA.
Furthermore, even if we do not get the same time reduction as with the \textit{direct} approach, it remains significant, halving the execution time or more.

Lastly, it can be seen that the \textit{indirect} PCA approach yields better estimators when we set \textit{n} = 8 instead of 17, as well as compared to the case in which PCA is not involved at all.
On the other hand, Accuracy\textsubscript{GT} is generally better when \textit{n} = 17, indicating that the higher the PC considered, the better the degree of generalization of the obtained theories.
The better results in ground-truth scores suggest that by reducing the dataset's complexity through \textit{indirect} PCA, ILASP can concentrate on the neural network's most salient predictive information. In fact, PCA is specifically designed to preserve maximal variance, thereby filtering out noise and redundant features. This ensure that the subsequent rule-learning phase operates on a distilled representation of the model's behavior. As a result, ILASP generates rules that capture the network's core decision criteria without being distracted by spurious correlations or minor fluctations in less significant feature. Thus, the outcome is a more accurate explanatory theory that aligns closely with the true user preferences reflected in the ground-truth data.
{All this evidence allows us to conclude that PCA-based approaches not only enable us to address \textbf{Q3}, but also contribute to providing more robust results for \textbf{Q2}.}
\paragraph{ILASP as classifier and PCA hypothesis.} The results in Table \ref{table:resultsClassificator} show the outcomes achieved by using ILASP directly on the user-labeled preferences in the dataset rather than using neural network predictions.
Note that in this case, the training set has 157 pairs, while the test set only 53 (since we have 210 couples for each user in the preferences' dataset).
As can be seen, even with \textit{indirect} PCA approach with \textit{n} = 8 we obtain an execution time that is significantly greater of the case in which we do not use PCA at all when using ILASP as global approximator on dataset with 190 pairs.
On the other hand, the \textit{direct} approach has an execution time comparable to that of the global approximation case, and generally provides better estimators.
Specifically, when using PCA \textit{directly} the increase of results do not stop at 10PC as before, but until 20.
These results support the two hypothesis regarding the use of \textit{direct} PCA approach: in this case, there is not the simplification induced by the approximation made by neural network, thus the feature space given to ILASP is more complex than before. 
This reflects in the absence of overfitting in \textit{direct} approach.
Furthermore, even if the \textit{indirect} PCA execution time increase with respect to the global case, the \textit{direct} PCA approach is not, strengthening the idea that its execution time reduction is given by the orthonormal basis created by PCA.

\begin{table}[h!]
    \centering
    \caption{ILASP as global approximator on training set of 45 pairs}
    {\tablefont
    \begin{tabularx}{\textwidth}{@{\extracolsep{\fill}} ccccccccc}
        \topline
        & Fidelity & Precision$_{BB}$ & Recall$_{BB}$ & Time(s) & \#WC & accuracy$_{GT}$ & precision$_{GT}$ & recall$_{GT}$\\
        \topline
        No PCA & 66.60\% & 37.74\% & 42.62\% & 72.27 & 3.5 & 58.79\% & 50.04\% & 48.56\%
        \midline
        \multirow{2}{*}{\makecell{Indirect \\(8PC)}} & 68.19\% & \textbf{45.49\%} & 42.45\% & 63.73 & \textbf{3.8} & \textbf{70.98\%} & \textbf{64.09\%} & 51.78\%\\
        & (+1.59\%) & \textbf{(+7.75\%)} & (-0.17\%) & (-8.54) & \textbf{(+0.3)} & \textbf{(+12.19\%)} & \textbf{(+14.05\%)} & (+3.22\%)
        \midline\textbf{}
        \multirow{2}{*}{\makecell{Indirect \\(17PC)}} & 67.68\% & 37.87\% & 37.86\% & 68.55 & 3.5 & 65.65\% & 59.99\% & \textbf{56.56\%}\\
        & (+1.08\%) & (+0.13\%) & (-4.76\%) & (-3.72) & (--) & (+6.86\%) & (+9.95\%) & \textbf{(+8.00\%)}
        \midline
        \multirow{2}{*}{\makecell{Direct \\(5PC)}} & \textbf{70.98\%} & 44.55\% & \textbf{49.25\%} & \textbf{1.83} & 1.3 & 51.49\% & 22.39\% & 26.35\%\\
        & \textbf{(+4.38\%)} & (+6.81\%) & \textbf{(+6.63\%)} & \textbf{(-70.44)} & (-2.2) & (-7.30\%) & (-27.65\%) & (-22.21\%)
        \midline
        \multirow{2}{*}{\makecell{Direct \\(10PC)}} & 70.22\% & 41.96\% & 41.99\% & 4.50 & 1.6 & 57.59\% & 28.09\% & 28.29\%\\
        & (+3.62\%) & (+4.22\%) & (-0.63\%) & (-67.77) & (-1.9) & (-1.20\%) & (-21.95\%) & (-20.27\%)
        \midline
        \multirow{2}{*}{\makecell{Direct \\(15PC)}} & 68.95\% & 41.96\% & 44.35\% & 6.62 & 1.4 & 58.98\% & 31.46\% & 34.17\%\\
        & (+2.35\%) & (+4.22\%) & (+1.73\%) & (-65.65) & (-2.1) & (+0.19\%) & (-18.58\%) & (-14.39\%)
        \midline
        \multirow{2}{*}{\makecell{Direct \\(20PC)}} & 69.02\% & 41.35\% & 40.49\% & 8.91 & 1.4 & 58.41\% & 31.05\% & 34.25\%\\
        & (+2.42\%) & (+3.61\%) & (-2.13\%) & (-63.36) & (-2.1) & (-0.38\%) & (-18.99\%) & (-14.31\%)
        \botline
    \end{tabularx}}
    \label{table:resultsGlobalData1}
\end{table}

\begin{table}[h!]
    \centering
    \caption{ILASP as global approximator on training set of 105 pairs}
    {\tablefont
    \begin{tabularx}{\textwidth}{@{\extracolsep{\fill}} ccccccccc}
        \topline
        & Fidelity & Precision$_{BB}$ & Recall$_{BB}$ & Time(s) & \#WC & accuracy$_{GT}$ & precision$_{GT}$ & recall$_{GT}$\\
        \topline
        No PCA & 71.62\% & 41.39\% & 42.11\% & 6102.15 & 5.0 & 57.90\% & 47.66\% & 35.27\%  
        \midline
        \multirow{2}{*}{\makecell{Indirect \\(8PC)}}
            & \textbf{73.65\%} & 42.26\% & 46.97\% & 970.54 & 4.9 & 60.95\% & \textbf{50.54\%} & 39.60\%\\
            & \textbf{(+2.03\%)} & (+0.87\%) & (+4.86\%) & (-5131.61) & (-0.1) & (+3.05\%) & \textbf{(+2.88\%)} & (+4.33\%)  
        \midline
        \multirow{2}{*}{\makecell{Indirect \\(17PC)}}
            & 71.81\% & 44.67\% & 42.97\% & 2617.40 & \textbf{5.0} & \textbf{61.71\%} & 46.84\% & \textbf{40.70\%}\\
            & (+0.19\%) & (+3.28\%) & (+0.86\%) & (-3484.75) & \textbf{(--)} & \textbf{(+3.81\%)} & (-0.82\%) & \textbf{(+5.43\%)}  
        \midline
        \multirow{2}{*}{\makecell{Direct \\(5PC)}}
            & 71.81\% & 44.49\% & 48.89\% & \textbf{3.67} & 1.4 & 50.29\% & 19.64\% & 27.64\%\\
            & (+0.19\%) & (+3.10\%) & (+6.78\%) & \textbf{(-6098.48)} & (-3.6) & (-7.61\%) & (-28.02\%) & (-7.63\%)  
        \midline
        \multirow{2}{*}{\makecell{Direct \\(10PC)}}
            & 72.44\% & 45.12\% & \textbf{49.14\%} & 12.83 & 1.6 & 57.21\% & 23.61\% & 27.61\%\\
            & (+0.82\%) & (+3.73\%) & \textbf{(+7.03\%)} & (-6089.32) & (-3.4) & (-0.69\%) & (-24.05\%) & (-7.66\%)  
        \midline
        \multirow{2}{*}{\makecell{Direct \\(15PC)}}
            & 70.22\% & 43.35\% & 46.97\% & 23.88 & 1.5 & 59.43\% & 26.18\% & 31.05\%\\
            & (-1.40\%) & (+1.96\%) & (+4.86\%) & (-6078.27) & (-3.5) & (+1.53\%) & (-21.48\%) & (-4.22\%)  
        \midline
        \multirow{2}{*}{\makecell{Direct \\(20PC)}}
            & 71.11\% & \textbf{47.63\%} & 43.61\% & 30.85 & 1.5 & 58.29\% & 24.89\% & 29.85\%\\
            & (-0.51\%) & \textbf{(+6.24\%)} & (+1.50\%) & (-6071.30) & (-3.5) & (+0.39\%) & (-22.77\%) & (-5.42\%)  
        \botline
    \end{tabularx}}
    \label{table:resultsGlobalData2}
\end{table}

\begin{table}[h!]
    \centering
    \caption{ILASP as global approximator on training set of 190 pairs}
    {\tablefont
    \begin{tabularx}{\textwidth}{@{\extracolsep{\fill}} ccccccccc}
        \topline
        & Fidelity & Precision$_{BB}$ & Recall$_{BB}$ & Time(s) & \#WC & accuracy$_{GT}$ & precision$_{GT}$ & recall$_{GT}$\\
        \topline
        No PCA & 64.95\% & 35.81\% & 35.47\% & 22210.65 & 5.0 & 65.78\% & 44.93\% & 42.98\%  
        \midline
        \multirow{2}{*}{\makecell{Indirect \\(8PC)}}
            & 65.78\% & 38.00\% & 37.20\% & 6802.32 & \textbf{5.0} & 63.56\% & \textbf{54.70\%} & \textbf{49.53\%}\\
            & (+0.83\%) & (+2.19\%) & (+1.73\%) & (-15408.33) & \textbf{(--)} & (-2.22\%) & \textbf{(+9.77\%)} & \textbf{(+6.55\%)}  
        \midline
        \multirow{2}{*}{\makecell{Indirect \\(17PC)}}
            & 65.59\% & 37.31\% & 37.15\% & 12106.05 & \textbf{5.0} & \textbf{65.65\%} & 53.96\% & 44.86\%\\
            & (+0.64\%) & (+1.50\%) & (+1.68\%) & (-10104.60) & \textbf{(--)} & \textbf{(-0.13\%)} & (+9.03\%) & (+1.88\%)  
        \midline
        \multirow{2}{*}{\makecell{Direct \\(5PC)}}
            & 67.17\% & 37.79\% & 38.44\% & \textbf{8.74} & 1.6 & 52.13\% & 25.23\% & 30.38\%\\
            & (+2.22\%) & (+1.98\%) & (+2.97\%) & \textbf{(-22201.91)} & (-3.4) & (-13.65\%) & (-19.70\%) & (-12.60\%)  
        \midline
        \multirow{2}{*}{\makecell{Direct \\(10PC)}}
            & \textbf{69.33\%} & \textbf{42.03\%} & \textbf{40.32\%} & 25.03 & 1.8 & 59.49\% & 28.49\% & 32.13\%\\
            & \textbf{(+4.38\%)} & \textbf{(+6.22\%)} & \textbf{(+4.85\%)} & (-22185.62) & (-3.2) & (-6.29\%) & (-16.44\%) & (-10.85\%)  
        \midline
        \multirow{2}{*}{\makecell{Direct \\(15PC)}}
            & 66.92\% & 36.64\% & 36.87\% & 49.58 & 1.9 & 58.98\% & 25.89\% & 34.54\%\\
            & (+1.97\%) & (+0.83\%) & (+1.40\%) & (-22161.07) & (-3.1) & (-6.80\%) & (-19.04\%) & (-8.44\%)  
        \midline
        \multirow{2}{*}{\makecell{Direct \\(20PC)}}
            & 65.71\% & 35.44\% & 35.33\% & 60.06 & 1.9 & 63.68\% & 30.52\% & 36.52\%\\
            & (+0.76\%) & (-0.37\%) & (-0.14\%) & (-22150.59) & (-3.1) & (-2.10\%) & (-14.41\%) & (-6.46\%)  
        \botline
    \end{tabularx}}
    \label{table:resultsGlobalData3}
\end{table}

\begin{table}[h!]
    \centering
    \caption{ILASP as classifier on preferences' training set of 157 pairs}
    {\tablefont
    \begin{tabularx}{\textwidth}{@{\extracolsep{\fill}} ccccccccc}
        \topline
        & Accuracy & Precision & Recall & Time(s) & \#WC & accuracy$_{GT}$ & precision$_{GT}$ & recall$_{GT}$\\
        \topline
        \makecell{Indirect \\(8PC)} & \textbf{77.52\%} & \textbf{48.99\%} & \textbf{50.86\%} & 155306.17 & \textbf{5.0} & 60.13\% & \textbf{42.67\%} & \textbf{42.32\%}
        \midline
        \multirow{2}{*}{\makecell{Direct \\(5PC)}}
            & 71.44\% & 42.05\% & 36.96\% & \textbf{5.51} & 1.7 & 54.34\% & 26.54\% & 26.94\%\\
            & (-6.08\%) & (-6.94\%) & (-13.90\%) & \textbf{(-155300.66)} & (-3.3) & (-5.79\%) & (-16.13\%) & (-15.38\%)  
        \midline
        \multirow{2}{*}{\makecell{Direct \\(10PC)}}
            & 71.80\% & 42.24\% & 38.03\% & 24.69 & 1.9 & \textbf{62.64\%} & 29.30\% & 34.19\%\\
            & (-5.72\%) & (-6.75\%) & (-12.83\%) & (-155281.48) & (-3.1) & \textbf{(+2.51\%)} & (-13.37\%) & (-8.13\%)  
        \midline
        \multirow{2}{*}{\makecell{Direct \\(15PC)}}
            & 72.09\% & 42.24\% & 38.17\% & 40.45 & 1.9 & 59.87\% & 29.89\% & 32.99\%\\
            & (-5.43\%) & (-6.75\%) & (-12.69\%) & (-155265.72) & (-3.1) & (-0.26\%) & (-12.78\%) & (-9.33\%)  
        \midline
        \multirow{2}{*}{\makecell{Direct \\(20PC)}}
            & 72.41\% & 42.81\% & 37.79\% & 58.77 & 2.0 & 58.87\% & 25.77\% & 34.35\%\\
            & (-5.11\%) & (-6.18\%) & (-11.07\%) & (-155247.40) & (-3.0) & (-1.26\%) & (-16.90\%) & (-7.97\%)  
        \botline
    \end{tabularx}}
    \label{table:resultsClassificator}
\end{table}

\label{subsection:global-results}

\subsection{Local approximation}
In our experiment using ILASP for local approximation, we tested different standard deviation values for the Gaussian noise applied to the queries.
Such experiment concerns \textit{indirect} PCA approach with \textit{n} = 8, this because it is the case in which we obtain the best trade-off between estimators, execution time and complex theories in the global approximation.
In fact, as described in Section \ref{subsection:global-results}, \textit{direct} PCA approach generally has better estimators and execution time results, but worse ground-truth results and simpler theories. On the other hand, \textit{indirect PCA} halves the execution time, and still obtain an overall increase in all the estimators, which is more marked when \textit{n} = 8.
Before to start to show and comment results, must be reported that at average of 5-10\% couples of test set have empty theory as explanation returned by ILASP.
Such cases occurred when the training set mainly consisted of couples labelled with uncertainty (about 90-95\% of the couples in the training set).
ILASP seeks to minimize the length of a theory plus the cost of not covering examples, but due to the limitation in the local range, it struggles to define a theory that properly covers couples with uncertainty, leading to empty set as a result.
Note that, as fully described later, it is possible to obtain a training set with over 90\% of couple with the same label when there is a high degree of locality (for example, std = 0.001).
\paragraph{Locality degree and general results.} The results are reported in Tables \ref{table:resultsLocalData1}, \ref{table:resultsLocalData2} and \ref{table:resultsLocalData3}.
Any instances where the returned theory is empty have not been included in the table. This decision was made because it is unclear how to test an empty theory, and these cases only account for 5-10\% of the total.
By definition of Gaussian noise, the choice of standard deviation affects the grade of similarity of generated data respect to the original query, as well as the labels predicted by the Neural Network, which are then fed to ILASP to validate the learned theory as an approximation.
A lower standard deviation increases the likelihood that elements of the training set are labeled with the same value, making ILASP's task simpler. 
Conversely, a higher standard deviation presents ILASP with a more challenging task. 
This trend is reflected in the results: as the standard deviation decreases, there is an increase in fidelity, precision\textsubscript{BB}, and recall\textsubscript{BB}, and a decrease in execution time.
However, this pattern does not hold when the training set comprises 190 pairs, as the execution time explodes with standard deviation = 0.001. 
As this is an isolated case with no apparent justification for the explosion, it is treated as an outlier.
On the other hands, lesser is the standard deviation, lesser are the differences between generated data respect to the original.
This means that are also lesser the weak constraints that ILASP can extrapolate from a training set, leading to less complex theory (and so with lesser \#WC), as visible in the results.
Finally, note that results for std = 1 are omitted from Table \ref{table:resultsLocalData3} because using ILASP on all pairs of the full training set became prohibitively time-consuming.

Lastly, ground-truth score yields better results when std is 0.1, and worse at lowers std.
This could suggest that considering a locality degree too small can lead to overfitting.
For this reason, and since we consider it the best trade-off between estimators, execution time, and number of weak constraints, we choose a standard deviation of 0.1 for the next experiments using local approximator.

{Once more, the results presented in the tables, along with the supporting evidence, confirm positive answers to \textbf{Q1} and \textbf{Q2}, as will be further clarified later in this section.}
\paragraph{Global vs Local execution time.} {When standard deviation is 0.1, } training set has 190 pairs and \textit{indirect} PCA is used with \textit{n} = 8, {our method} lead to an average execution time of only 401.62 seconds for a single query.
This is significantly lower than the corresponding time of 6802.32 seconds for global approximation.
But when using ILASP as a global approximator, the returned theory can be used to predict the number of pairs needed, whereas for local approximation, the theory can only be used for the specific query ILASP was used on.
In the given example, local approximator need 401.62 seconds for each pair, while the test set used for global approximator has 105 pairs.
This means that the local approximator would take 42170.1 seconds to approximate all the pairs in the test set, compared to the 6802.32 seconds needed by the global approximator. 
However, the theories returned by the local approximator are more specific to individual pairs, resulting in significantly better fidelity, precision\textsubscript{BB}, and recall\textsubscript{BB} compared to the global approach.
In conclusion, {also considering \textbf{Q3},} the global approximator appears to be more efficient than the local one, while the local approach is more effective than the global one.
\paragraph{Local approximation performances.} {The result discussed in previous paragraph} is also visible in Table \ref{table:resultsLocalData4}, where are reported the result of ILASP as local approximator both when \textit{indirect} PCA approach is applied (with \textit{n} = 8, 17) and when is not, with standard deviation equal to 0.1.
We did not consider \textit{direct} approach because of the already too little complex theory, which would become further simplified by the use of this technique, as seen in global approximation study.
In terms of fidelity, precision\textsubscript{BB} and recall\textsubscript{BB}, these results behave similarly to what seen in the global case, but now the increase in significantly greater {, answering to \textbf{Q2} also from the local perspective}.
In fact, while in the global case the greater difference with the \textit{indirect} PCA was of 2.03\%, here we get a difference of about 6.20\% with \textit{n} = 8 and 14.29\% with \textit{n} = 17.
In our opinion, this significant difference is justified because local approximators generally seek less complex theories, requiring fewer features (e.g.: \textit{browning} preparation) to discriminate preferences among recipes. 
Thus, \textit{indirect} PCA improves results by effectively helping ILASP select necessary features and removing those with minor roles. 
Conversely, in the global case, ILASP seeks more complex theories where dimensionality reduction only marginally enhances performance, as even minor role features remain important for general decision-making.
Put simply, in the local case we get less complex but more accurate theories, which are better enhanced with respect to the global case (where we necessarily need more complex theories) by the dimensionality reduction.

\begin{table}[h!]
    \centering
    \caption{ILASP as local approximator on training set of 45 pairs}
    {\tablefont
    \begin{tabularx}{\textwidth}{@{\extracolsep{\fill}} ccccccccc}
        \topline
        std & Fidelity & Precision$_{BB}$ & Recall$_{BB}$ & Time(s) & \#WC & accuracy$_{GT}$ & precision$_{GT}$ & recall$_{GT}$\\
        \topline
        \makecell{1} & 64.59\% & 41.31\% & 36.43\% & 75.61 & \textbf{2.36} & 62.50\% & 49.28\% & 41.49\%
        \midline
        \multirow{2}{*}{\makecell{0.1}} 
            & 83.54\% & 63.40\% & 60.81\% & 46.30 & 1.33 & \textbf{63.20\%} & \textbf{52.87\%} & \textbf{45.83\%}\\
            & (+18.95\%) & (+22.09\%) & (+24.38\%) & (-29.31) & (-1.03) & \textbf{(+0.70\%)} & \textbf{(+3.59\%)} & \textbf{(+4.34\%)}
        \midline
        \multirow{2}{*}{\makecell{0.01}} 
            & 96.97\% & \textbf{71.61\%} & 63.40\% & 4.61 & 1.02 & 59.64\% & 52.40\% & 33.49\%\\
            & (+32.38\%) & \textbf{(+30.30\%)} & (+26.97\%) & (-71.00) & (-1.34) & (-2.86\%) & (+3.12\%) & (-8.00\%)
        \midline
        \multirow{2}{*}{\makecell{0.001}} 
            & \textbf{97.97\%} & 66.98\% & \textbf{68.22\%} & \textbf{4.26} & 1.02 & 58.83\% & 41.72\% & 31.84\%\\
            & \textbf{(+33.38\%)} & (+25.67\%) & \textbf{(+31.79\%)} & \textbf{(-71.35)} & (-1.34) & (-3.67\%) & (-7.56\%) & (-9.65\%)
        \botline
    \end{tabularx}}
    \label{table:resultsLocalData1}
\end{table}

\begin{table}[h!]
    \centering
    \caption{ILASP as local approximator on training set of 105 pairs}
    {\tablefont
    \begin{tabularx}{\textwidth}{@{\extracolsep{\fill}} ccccccccc}
        \topline
        std & Fidelity & Precision$_{BB}$ & Recall$_{BB}$ & Time(s) & \#WC & accuracy$_{GT}$ & precision$_{GT}$ & recall$_{GT}$\\
        \topline
        \makecell{1} & 65.82\% & 42.66\% & 36.60\% & 960.43 & \textbf{2.57} & 62.04\% & 45.30\% & 39.46\%
        \midline
        \multirow{2}{*}{\makecell{0.1}} 
            & 86.09\% & 66.01\% & 67.02\% & 192.16 & 1.43 & \textbf{64.05\%} & \textbf{56.56\%} & \textbf{46.09\%}\\
            & (+20.27\%) & (+23.35\%) & (+30.42\%) & (-768.27) & (-1.14) & \textbf{(+2.01\%)} & \textbf{(+11.26\%)} & \textbf{(+6.63\%)}
        \midline
        \multirow{2}{*}{\makecell{0.01}} 
            & 96.77\% & 66.84\% & 66.05\% & 4.47 & 1.06 & 58.81\% & 51.17\% & 33.17\%\\
            & (+30.95\%) & (+24.18\%) & (+29.45\%) & (-955.96) & (-1.51) & (-3.23\%) & (+5.87\%) & (-6.29\%)
        \midline
        \multirow{2}{*}{\makecell{0.001}} 
            & \textbf{98.06\%} & \textbf{68.32\%} & \textbf{68.43\%} & \textbf{3.89} & 1.00 & 58.09\% & 37.81\% & 29.49\%\\
            & \textbf{(+32.24\%)} & \textbf{(+25.66\%)} & \textbf{(+31.83\%)} & \textbf{(-956.54)} & (-1.57) & (-3.95\%) & (-7.49\%) & (-9.97\%)
        \botline
    \end{tabularx}}
    \label{table:resultsLocalData2}
\end{table}

\begin{table}[h!]
    \centering
    \caption{ILASP as local approximator on training set of 190 pairs}
    {\tablefont
    \begin{tabularx}{\textwidth}{@{\extracolsep{\fill}} ccccccccc}
        \topline
        std & Fidelity & Precision$_{BB}$ & Recall$_{BB}$ & Time(s) & \#WC & accuracy$_{GT}$ & precision$_{GT}$ & recall$_{GT}$\\
        \topline
        \makecell{1} & -- & -- & -- & -- & -- & -- & -- & --
        \midline
        \makecell{0.1} & 84.40\% & 62.10\% & 60.40\% & 401.62 & \textbf{1.48} & \textbf{61.23\%} & 52.10\% & \textbf{43.11\%}
        \midline
        \multirow{2}{*}{\makecell{0.01}}
            & 96.53\% & \textbf{68.88\%} & 62.10\% & \textbf{7.81} & 1.07 & 59.53\% & \textbf{54.53\%} & 34.95\%\\
            & (+12.13\%) & \textbf{(+6.78\%)} & (+1.70\%) & \textbf{(-393.81)} & (-0.41) & (-1.70\%) & \textbf{(+2.43\%)} & (-8.16\%)  
        \midline
        \multirow{2}{*}{\makecell{0.001}}
            & \textbf{98.03\%} & 65.26\% & \textbf{65.02\%} & 1101.55 & 1.02 & 59.65\% & 46.11\% & 31.83\%\\
            & \textbf{(+13.63\%)} & (+3.16\%) & \textbf{(+4.62\%)} & (+699.93) & (-0.46) & (-1.58\%) & (-5.99\%) & (-1.28\%)  
        \botline
    \end{tabularx}}
    \label{table:resultsLocalData3}
\end{table}

\begin{table}[h!]
    \centering
    \caption{ILASP as local approximator on training set of 105 pairs (std = 0.1)}
    {\tablefont
    \begin{tabularx}{\textwidth}{@{\extracolsep{\fill}} ccccccccc}
        \topline
        & Fidelity & Precision$_{BB}$ & Recall$_{BB}$ & Time(s) & \#WC & accuracy$_{GT}$ & precision$_{GT}$ & recall$_{GT}$\\
        \topline
        \makecell{No PCA} & 79.89\% & 62.62\% & 60.52\% & 216.17 & 1.60 & 64.60\% & 51.96\% & 46.80\%
        \midline
        \multirow{2}{*}{\makecell{Indirect \\(8PC)}}
            & 86.09\% & \textbf{66.01\%} & 67.02\% & \textbf{192.16} & 1.43 & \textbf{64.05\%} & \textbf{56.56\%} & 46.09\%\\
            & (+6.20\%) & \textbf{(+3.39\%)} & (+6.50\%) & \textbf{(-24.01)} & (-0.17) & \textbf{(-0.55\%)} & \textbf{(+4.60\%)} & (-0.71\%)  
        \midline
        \multirow{2}{*}{\makecell{Indirect \\(17PC)}}
            & \textbf{94.18\%} & 65.84\% & \textbf{69.18\%} & 318.04 & \textbf{1.49} & 63.76\% & 51.17\% & \textbf{46.72\%}\\
            & \textbf{(+14.29\%)} & (+3.22\%) & \textbf{(+8.66\%)} & (+101.87) & \textbf{(-0.11)} & (-0.84\%) & (-0.79\%) & \textbf{(+0.92\%)}  
        \botline
    \end{tabularx}}
    \label{table:resultsLocalData4}
\end{table}

\label{subsection:local-results}

\section{Conclusion and future developments}
In recent years, XAI has emerged as a critical discipline to make machine learning models transparent and interpretable, delivering tangible benefits in domains such as healthcare, finance and recommendation systems. In the specific context of preference learning, the ability to explain a black-box model’s rankings not only help end users to understand its decision, but also enables bias detection and performance improvement through an iterative feedback cycle. In this context, a promising approach in this field is that of Inductive Logic Programming, a subfield of artificial intelligence which aims to generate intuitive and readable explanation of the decision-making of black box, which can be fed even to lay users.
In this work we propose several approaches for using of ILP to explain black-box preference learning system while dealing with the issue of time complexity.
In fact, among the challenges with XAI methods~\citep{Saeed2023XAIchallenges}, Scalability is one of the most pressing ones, caused by increasingly dimensionality in data or cases for which huge number of explanation are needed.
Specifically, we propose ILASP as global and local approximator of a Neural Network involving PCA in the process to reduce the dimensionality of the dataset and experimenting by sampling dataset of increasing size from dataset feature space.
This not only led to a significant reduction in execution time, but also helps to improve results, thanks to the ability of PCA to compress great part of the informative content of the dataset in the firsts Principal Components.

Within this work, we also created a food preference dataset, which is at the base of the experiments.
This dataset is based on a survey where participants ranked Italian recipes.
The results of the survey are then processed in order to obtain the data used to encode the ranking problem as pairwise comparison, including the uncertainty scenario.
Another challenge of XAI in which many efforts are invested, is to obtain methods with better fairness. In fact, a given explanation that entail a prediction might not be the correct one anyway, as the user could have another reason. One way to turn-around these problems is the use of ground-truth to validate the results. To this end, we defined ground-truth scores to evaluate theories generated by ILASP. In this way, we can assess ILASP performances as approximator using standard metric in XAI like fidelity~\citep{gui18survey}, and as faithful representations of user's actual preferences with aforementioned ground-truth scores.

As mentioned in our paper, a key strength of ILASP lies in its ability to produce theories that can be readily rendered into natural language. In future work, we plan to enhance this capability leveraging either ad-hoc parser or prompt-engineered large language models to translate ILASP’s output. This will enable lay end users to directly assess both the accuracy and the interpretability of the generated theories.

The resulting dataset is unbalanced on uncertainty class, and this reflected is visible in the final results.
In future works, we aim to address this issue.
An example to deal with it is by considering sampling techniques during training and test sets definition both for global and local experiments (e.g. by taking into account only sets of couple which labels given by the Neural Network are balanced in the set).
Furthermore, as discussed in Section \ref{subsection:hyperparameters}, sampled dataset with 190 pairs used in global approximation is actually less similar to the dataset used to train the black-box model than the sampled dataset with 105 pairs. This lead ILASP to obtain worst results in the sampled dataset with 190 pairs, despite its size. In future work, we plan to incorporate similarity metrics among dataset, e.g., Maximum Mean Discrepancy~\citep{gretton2012MMD}, to ensure that our sampled datasets closely resemble the black‐box training distribution. This should enable ILASP to achieve better results when used on larger samples.

As mentioned, we involved PCA during experiments to deal with the time execution problem that a system like ILASP brings out when used on high-dimensional dataset.
Specifically, we proposed two approaches, which are \textit{indirect} and \textit{direct} PCA.
While the first obtain more correct (with respect to the ground-truth) and complex theories, the second obtains the better results in terms of estimators.
Given that, in both cases, we obtain important reduction of execution time, passing from execution time of about 6 hours, to less than 1 minute.
A possible development that is worth to be further investigated could be the composition of these two approaches.
In fact, \textit{direct} approach could be used as a sort of pre-processing for \textit{indirect} one.
Instead of considering the first \textit{j} PCs in (\ref{eq:indirectlyPCA}) for the definition of features to maintain, we could use the PCs contained in the theory returned by the \textit{direct} approach.
In turn, we can use ILASP on the same pairs using the resulting features and the corresponding theory as a starting point.
Such composition should exploit the advantages of both \textit{direct} and \textit{indirect} approaches, leading to better performance without oversimplified theories.

As fully described in the previous chapter, ILASP appears to be more efficient when used as a global approximator, while it is more effective as a local approximator.
Although local approximation yields good results compared to the global case, it also has a longer execution time because it can only be used for a pair, whereas the global approximator can be used for a set of pairs.
We anticipate that transfer learning could lead to improvements in this regard.
For instance, one could use the theory generated by the global approximator as a basis for the local approximator to reduce execution time. 
Besides that, other approaches could be investigated in the future in order to obtain better results, such as neurosymbolic methods.
{Furthermore, we note that this study is restricted to structured, symbolic inputs: extending the approach to raw, non-symbolic modaliteis (e.g. images, audio or raw text) would require additional concept-extraction or representation-learning steps (following a neurosymbolic approach) and is therefore left for future exploration.}
Finally, integrating diverse data such as gender, age, and region of origin into the existing ILASP framework may lead to better results, especially in a scenario like preference learning.

\label{section:conclusions}

\appendix

\section{ILASP technical details}
\label{appendixA}
{In Section~\ref{par:ILASP}, we discussed the main components of ILASP, focusing primarily on their theoretical aspects rather than on technical details.
This appendix provides the complete syntactic specification and additional technical details omitted from the main text, allowing the reader to gain a full understanding of the methodologies employed in this work.}
\paragraph{Mode declaration and Placeholder.} {Placeholder are terms of the form $var(t)$ or $const(t)$ for some constant term $t$ expressing that such terms can be replaced by any variable or constant of type $t$.
Each constant $c$ which is allowed to replace a $const$ term of type $t$ is specified as $\#constant(t, c)$.
So, an atom $a$ is compatible with a mode declaration $M$ if each of the placeholder constants and placeholder variables in $M$ has been replaced by a constant or variable of the correct type.
For instance $\#modeo(\cdot)$ declarations (also known as \textit{optimization body declaration}) forces ILASP to include weak constraints of the provided form in the search space.
There exist five types of mode declaration, which are:
\begin{itemize}\setlength\itemsep{0.5em}
    \item $\#modeh(p(\cdot))$ to declare what can be in the head ($\#modeha(p(\cdot))$ if we want to allow aggregates).
    \item $\#modeb(p(\cdot))$ to declare what can be in the body.
    \item $\#modec(\cdot)$ to declare conditions, where the argument can be, for instance $var(t1) > var(t2)$.
    \item $\#modeo(\cdot)$ declare what atoms can appear in a weak constraint.
\end{itemize}
\noindent For each mode declaration can be defined the \textit{recall}, which is a positive integer value specifying the maximum number of times that the mode declaration can be used in each rule.
For example,$\#modeb(2, p(var(t)))$ indicate that the atom $p(|\cdot)$, which take as argument a variable of type $t$, can be used at most two times.}

{Finally, consider the listing 1 of Example 2:}
\setcounter{lstlisting}{0}
\begin{lstlisting}[caption=constraints generated by the first line in Example 2]
    :~ value(p,V1).[1@2, V1]
    :~ value(p,V1).[-1@2, V1]
    :~ value(p,V1).[V1@2, V1]
    :~ value(p,V1).[-V1@2, V1]
    :~ value(p,V1).[V1@1, V1]
    :~ value(p,V1).[-1@1, V1]
    :~ value(p,V1).[1@1, V1]
    :~ value(p,V1).[-V1@1, V1]
\end{lstlisting}
\noindent {Notably, the} \verb|constant| \noindent {of type} \verb|val| \noindent {is defined as} \verb|p| \noindent {using }\verb|#constant(val, p)| \noindent {below, and the possible \textit{weights} for the constraints are $1$ and $-1$ times the value of the variable in the constraint, as specified through the} \verb|#weight| \noindent {declaration.}
\paragraph{Positive and Negative Examples.} {Syntactically, a positive example is defined as follows:}
\begin{equation}
    \verb|#pos(id, i1, ..., im, e1, ..., en).|
\end{equation}
\noindent {where} \verb|id| {identifies the example, $E^{inc}$ = }\verb|i1|, ... \verb|in| {are atoms called \textit{inclusions} while $E^{exc}$ = }\verb|e1|, ..., \verb|en| {atoms called \textit{exclusions}, and serve to define what atoms should be included and excluded by an answer set which ``cover'' that positive example.
Let define as $AS(P)$ the set of all answer sets of a ASP program $P$, then a positive example is covered by an answer set if there exist an answer set $A \in AS(P)$ such that $(E^{inc} \subseteq A) \wedge (E^{exc} \cap A = \emptyset)$ (in this case we say that $A$ extends $E= \langle E^{inc}, E^{exc} \rangle$).
Negative examples can be defined similarly with the only difference that a negative example is covered if there is no answer set of the learned program that extends it.}
\paragraph{ordering examples.} {There exist two kind of ordering example, namely \textit{cautious ordering} and \textit{brave ordering}. Syntactically we can define them as follow:}
\begin{equation}
    \begin{aligned}
        \verb|#cautious_ordering(e1, e2, s)|\\
        \verb|#brave_ordering(e1, e2, s)|
    \end{aligned}
\end{equation}
{Where} \verb|e1| {and} \verb|e2| {are two examples and} \verb|s| {denoted one of the rational operator $=, \ge, \le, >, <$.}
{If} \verb|s| {is equal to $<$, \textit{cautious ordering} states that every answer set which extends} \verb|e1| {must be preferred to any answer set which extends} \verb|e2| {(if} \verb|s| {is equal to $>$, it means the opposite, while if it assumes $=$ value, means that they are equally preferred).
\textit{Brave orderings}, considering} \verb|s| {is equal to $<$, state that at least one answer set which extends} \verb|e1| {must be preferred to any answer set which extends} \verb|e2| {(if} \verb|s| {is equal to $>$, it means the opposite, while if it assumes $=$ value, means that they are equally preferred).
}

\section{Neural Network training and architecture} \label{AppendixB}
In this section, we describe the experimented NN, describing its training and architecture.

The experiments follow an inter-users approach, and so for each user is trained a neural network with the same structure but with the user's preferences as input.
As described in Section \ref{subsection:survey} the problem that these models have to solve is a ternary classification problem.
The samples are pairs of recipes, while the classes are the preference relationships that occur among the recipes with respect to the user.
Specifically, a sample has the form $\langle ID1, ID2\rangle = [a_1, a_2, ..., a_k, b_1, b_2, ..., b_k]$, where $a_i$ and $b_i$ are the feature associated to recipes with ID1 and ID2, respectively.
On the other hand, classes are labeled with 1 (``the first recipe is preferred over the second''), -1 (``the second recipe is preferred over the first'') and 0 (``it is uncertain which recipe is preferred between the two'').
Before training, PCA is applied to the data for dimensionality reduction following the ``Kaiser rule'', that is, taking into account the PCs with corresponding eigenvalue greater or equal to 1.
Thanks to the PCA we reduced each recipe to have 17 feature, so each sample $\langle ID1, ID2\rangle$ has 34 feature.
Note that differently from ILASP, data are feed considering ingredient classes instead meta-classes.
Finally, for each user we have 210 pairs, which are divided in 157 pairs for the training set and 53 for the testing set.

To define the architecture of the NN to use for the experiment we perform a fine-tuning of hyperparameters.
Tuned hyperparameters are activation function, optimization function, number of nodes for hidden layers (32, 64, and 128), the learning rate (0.01, 0.001, 0.0005, 0.0001 and 0.0005) and the use or not of dropout (if used we consider 0.1, 0.2 and 0.5 as dropout value) and batch normalization.
Because of the huge number of combination with such hyperparameters to tune, we make a two-stages tuning.
In the first stage we tuned activation functions (setting a constraint to avoid using linear activation function for more than one hidden layer) and the optimization function to use, fixing the learning rate to 0.0005, the number of nodes to 64 and the use of dropout (with 0.1 dropout value) and batch normalization, while in the second stage we tuned the remaining fixing activations and optimization function to those found in the first stage.

The architecture which return the best results, that are $82.72\%$ of accuracy, $83.26\%$ of precision and $82.43\%$ of recall is reported in Table \ref{table:neural_network}. Finally, in~\citep{fossemò2022inductive} we conducted comparative study on this model and tuned Support Vector Machines and K-nearest Neighbors, obtaining best results with Neural Network.

\begin{table}[h!]
    \centering
    \caption{Deep neural network obtained after the tuning of hyperparameters}
    {\tablefont\begin{tabularx}{0.87\textwidth}{@{\extracolsep{\fill}} ccc}
        \topline
         & Activation function & Nodes \\
        \topline
        Input layer & tanh & 64
        \midline
        Dropout & \multicolumn{2}{c}{Rate al 10\%}
        \midline
        Hidden layer & ReLU & 64
        \midline
        Dropout & \multicolumn{2}{c}{Rate al 10\%}
        \midline
         & \multicolumn{2}{c}{batch normalization}
        \midline
        Hidden layer & linear & 64
        \midline
        Dropout & \multicolumn{2}{c}{Rate al 10\%}
        \midline
        Output layer & softmax & 64 \\
        \topline
        Optimization function & \multicolumn{2}{c}{SGD}
        \midline
        learning rate & \multicolumn{2}{c}{0,0005}
        \botline
    \end{tabularx}}
	\label{table:neural_network}
\end{table}

\newpage

\section{Ingredient classes vs meta-classes}
In table \ref{table:ingredients-classes} are reported the ingredient meta-classes and corresponding classes.

\label{AppendixC}
\begin{table}[h!]
	\centering
    \caption{Recipes Dataset features}
    {\tablefont\begin{tabularx}{\textwidth}{@{\extracolsep{\fill}} cp{0.75\textwidth}}
        \topline
        \textbf{Ingredient meta-class} & \makecell{\textbf{Ingredient classes}} \\ 
        \topline
        \textbf{Cereals} & \makecell{Cereals}
        \midline
        \textbf{Dairies} & \makecell{Butter; Cheese; Milk and cream; Soft cheese; Yogurt}
        \midline
        \textbf{Eggs} & \makecell{Eggs}
        \midline
        \textbf{Flouries} & \makecell{Bread; Flour}
        \midline
        \textbf{Fruit} & \makecell{Berries; Citrus fruits; Nuts; Salty fruits; Sweet fruits}
        \midline
        \textbf{Herb, Spices, Seasonings} & \makecell{Broth; Condiments; Herbs; Spices; Wines and Spirits}
        \midline
        \textbf{Meat} & \makecell{Beef; Chicken meat; Pork meat; Rabbit meat; Sheep meat; \\Turkey Meat}
        \midline
        \textbf{Mushrooms and Truffles} & \makecell{Mushrooms}
        \midline
        \textbf{Pasta} & \makecell{Dry pasta; Fresh pasta}
        \midline
        \textbf{Seafood} & \makecell{Lake fish; Molluscs; Sea fish; Shellfish}
        \midline
        \textbf{Sweeteners} & \makecell{Sweeteners}
        \midline
        \textbf{Vegetables} & \makecell{Greens; Legumes; Vegetables}
        \botline
    \end{tabularx}}
    \label{table:ingredients-classes}
\end{table}

\section{ILASP code structure to approximate the Neural Network}
\label{AppendixD}
In this section we discuss the code structure of the ILASP program to obtain a theory approximating the neural network in our context.

\begin{lstlisting}[caption=Code structure to apply ILASP to obtain a theory approximating the Neural Network]
    #pos(item-i, {}, {}, {category(v1). value(cost, v2). value(difficulty, 
    v3). value(prepTime, v4). value(cereals, v5). ... value(vegetables, 
    v16). value(boiling, v17). ... value(stewing, v24).}).

    #maxv(1).
    #maxp(h).

    #modeo(1, value(const(val), var(val))).
    #modeo(1, category(const(mg), (positive)).
    #weight(val).
    #weight(1).
    #weight(-1).
    #constant(val, category).
    #constant(val, cost).
    #constant(val, difficulty).
    #constant(val, prepTime).
    #constant(mg, 1).
    ...
    #constant(mg, 5).
    #constant(val, cereals).
    ...
    #constant(val, vegetables).
    #constant(val, boiling).
    ...
    #constant(val, stewing).
    #brave_ordering(o_n@p_n, item-o_j1, item-o_j2, s_n).
\end{lstlisting}

\noindent where $v_1, ..., v_{24} \in \mathbb{N}$; $h \in \{1, ..., 5\}$; \verb|item-i| denote the $i_{th}$ recipe of the dataset, with $1 \le i \le |D|$ (where $|D|$ denote the dataset size); \verb|o_n| denote the $n_{th}$ brave ordering in the program and \verb|p_n| the corresponding associated penalty, with $1 \le n \le N$. With \verb|item-o_j1| (\verb|item-o_j2|) we identify the first (second) item present in the $n_{th}$ ordering example. Finally, with \verb|s_n| we denote the symbol of the corresponding brave ordering. Clearly $|D| = 100$ (since our recipes dataset has 100 recipes), while $N$ varies depending on the experiments. Furthermore, since we are in the global approximation case, $p_n = 1\ \forall\ n = 1, ..., N$. Note that, since we want to obtain a theory composed of weak constraints, we use \verb|brave_ordering| and \verb|#modeo(|$\cdot$\verb|)| as mode declaration. With \verb|brave_ordering| we are inducing a preference relation between answer sets, depending on the value of \verb|s_n|. Here the pairs (\verb|item-o_j1|, \verb|item-o_j2|) correspond to the $N$ sampled pair ($j_1, j_2$) (we refer to the $n_th$ sampled pair with the notation ($j_1, j_2$)$_n$), while \verb|s_j| is defined as follow:

\begin{equation}
    \verb|s_n| = 
    \begin{cases}
        <\;\; if\;\; B((j_1, j_2)_n) = 1\\
        =\;\; if\;\; B((j_1, j_2)_n) = 0\\
        >\;\; if\;\; B((j_1, j_2)_n) = -1\\
    \end{cases}
    \label{eq:s_value}
\end{equation}

\noindent note that here ``$<$'' should not be confused with `` \verb|item-o_j2| is preferred over \verb|item-o_j1| '', since in ILASP it means the opposite, that is `` \verb|item-o_j1| is preferred over \verb|item-o_j2|''. This difference comes from the cost concept introduced by ASP, explained in Section \ref{subsection:ASP}. Finally, \verb|#modeo(|$\cdot$\verb|)| mode declaration serves to give shape to our weak constraint, which will presents as the one reported below as an example
\setcounter{example}{2}
\begin{example}
\leavevmode\vspace{0ex}
\begin{verbatim}
    :~value(vegetables, V1).[-V1@1, V1]
    :~value(meat, V1).[-V1@2, V1]
    :~value(difficulty, V1).[-V1@3, V1]
    :~value(stewing, V1).[V1@4, V1]
    :~value(dairies, V1), category(3).[V1@5, V1]
\end{verbatim}

\end{example}

\noindent note that this is a theory obtained with $h = 5$ and that the presence of \verb|V1| as weight denotes that the penalty cost paid at the corresponding level of priority (if weak constraint is activated) is equal to the value of \verb|V1| in the atoms in which it appears.

\end{document}